\pdfoutput=1
\documentclass[10pt,letterpaper]{article}
\usepackage[pagenumbers]{style}
\usepackage{makecell}
\usepackage{multirow}
\usepackage[T1]{fontenc}

\usepackage[hypertexnames=false,pagebackref,breaklinks,colorlinks,allcolors=reporttheme]{hyperref}
\newcommand{\benchmarkname}{DexWM-Bench}
\title{\benchmarkname{}: Evaluating Fine-Grained Control and Interaction Fidelity for Dexterous World Models}

\author[1,2]{Xinyi Chen}
\author[3,2]{Bin Wang}
\author[4]{Zhongxi Chen}
\author[2]{Haotian Liang}
\author[2]{Kailin Li}
\author[5]{Zhixuan Liang}
\author[4,2]{Yao Mu}

\affiliation[1]{Fudan University}
\affiliation[2]{Shanghai Artificial Intelligence Laboratory}
\affiliation[3]{Zhejiang University}
\affiliation[4]{Shanghai Jiao Tong University}
\affiliation[5]{University of Hong Kong}

\begin{document}
\maketitle
\begin{abstract}
World models for dexterous manipulation promise to support planning and policy learning by predicting the consequences of robot actions.
Their utility depends on faithfully capturing fine-grained finger motions and hand-object interactions, where small control differences can change manipulation outcomes. However, existing evaluation methods provide limited systematic coverage of these dynamics.
To address this gap, we introduce \textbf{\benchmarkname{}}, a dedicated benchmark that evaluates dexterous world models along three complementary dimensions, namely \textbf{fine-grained control accuracy}, \textbf{interaction fidelity}, and \textbf{downstream execution utility}.
It combines full-task sequences, critical interaction clips covering both successful and failed outcomes, and controlled action comparisons that isolate the motion of a targeted arm or finger.
These evaluations examine whether commanded motions are realized, whether object responses and interaction outcomes match the reference, and whether generated videos support task execution when decoded into robot actions.
We further evaluate generalization across appearance changes, object replacements, and novel tasks.
We evaluate multiple widely-used world models spanning general-purpose video generators, action-conditioned world models, and world action models under the same protocol.
Our results motivate action-conditioning mechanisms tailored to dexterous hands and persistent modeling of object states through occlusion.
\end{abstract}

\section{Introduction}
\label{sec:intro}

\begin{figure}[t]
    \centering
    \includegraphics[width=1\textwidth]{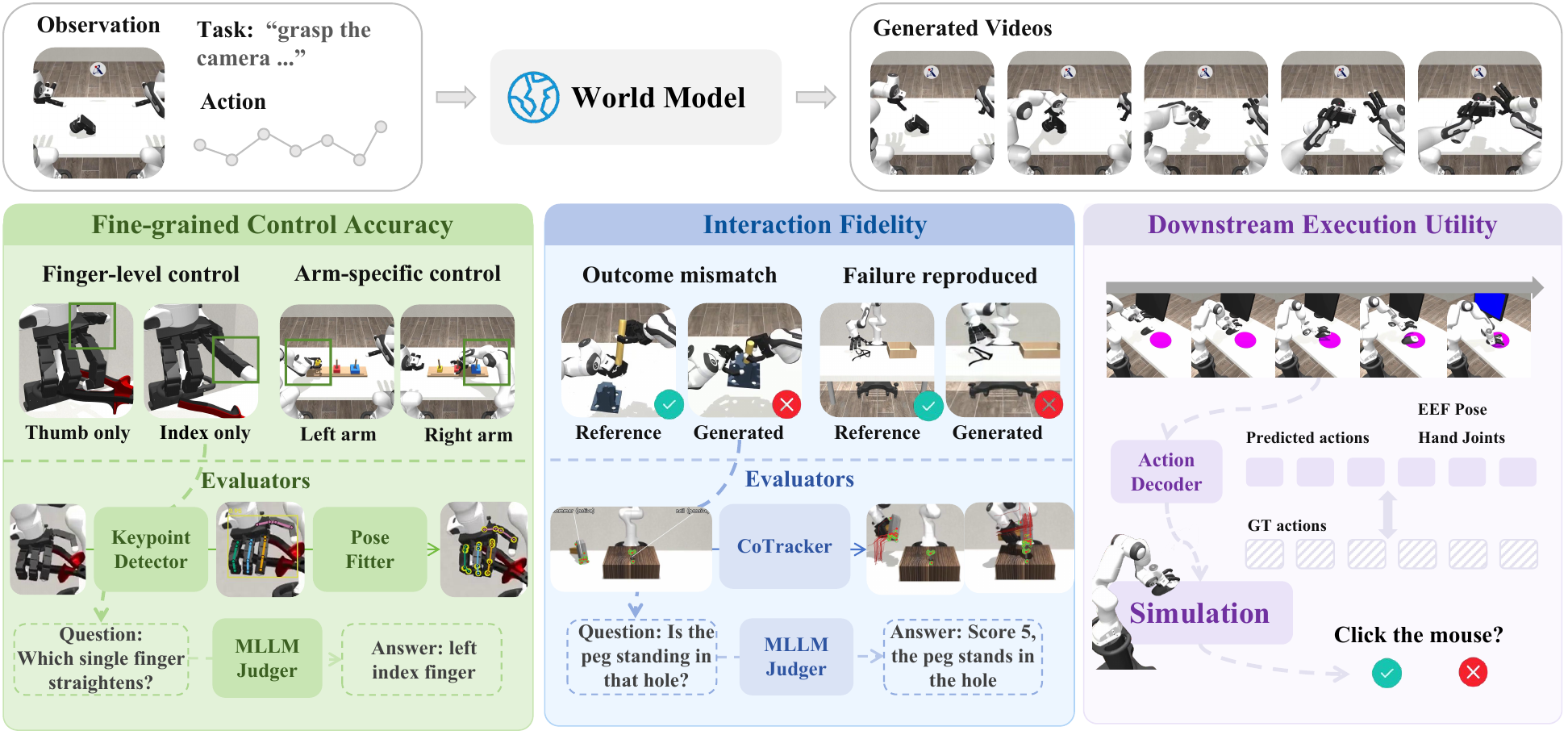}
    \caption{\textbf{Overview of \benchmarkname{}.} Conditioned on the initial observation, task instruction, and robot actions, a dexterous world model generates future videos. \benchmarkname{} evaluates these videos along three dimensions. \textbf{Fine-grained control accuracy} probes whether commanded motions appear only in the intended finger or arm. \textbf{Interaction fidelity} checks whether object trajectories and interaction outcomes, including failures, match the reference. \textbf{Downstream execution utility} pairs imagined video plans with an action decoder and evaluates task completion in simulation and agreement with reference actions.}
    \label{fig:teasor}
\end{figure}

World models for dexterous manipulation have recently attracted growing attention as learned simulators to anticipate the consequences of dexterous robot actions.
Conditioned on robot actions, hand motions, and task instructions, these models predict future visual observations or environment states to support planning, policy evaluation, and policy improvement.
Prior work has explored dexterous manipulation planning driven by future-state prediction \citep{dexwm}, video simulation with fine-grained action controllability \citep{mask2realwm}, and closed-loop policy learning that couples state prediction with outcome evaluation \citep{motus2}.
Compared with parallel-jaw manipulation, dexterous manipulation involves higher-dimensional control and more complex hand-object interaction, where small deviations in local motion can alter object responses and determine whether an interaction succeeds.
Frequent occlusion of objects by the hand requires models to track object states even when objects are partially or fully hidden.

Existing benchmarks evaluate visual quality, physical executability, and downstream utility~\citep{worldarena,robowmbench,kinebench,h2rbench}, but offer limited assessment of independent finger control and the resulting object interactions.
Coordinated finger movements in demonstrations can conceal whether a model responds to individual commands. In generated videos, objects may reappear with an incorrect shape or appearance after hand occlusion, even when the video remains visually coherent.
Evaluating dexterous world models therefore requires targeted tests of local action responses and their consequences.

To address this gap, we introduce \textbf{\benchmarkname{}}, a benchmark for evaluating fine-grained control and interaction fidelity in dexterous world models.
Built on DexJoCo~\citep{dexjoco}, it covers grasping, insertion, tool use, and bimanual manipulation through full-task sequences, critical interaction clips containing successful and failed outcomes, and controlled action comparisons.

\benchmarkname{} evaluates models along three dimensions.
\textbf{Fine-grained control accuracy} assesses hand geometry and responses to targeted arm or finger commands.
\textbf{Interaction fidelity} measures agreement with reference object trajectories and interaction outcomes.
\textbf{Downstream execution utility} examines whether actions decoded from generated videos support task completion from selected intermediate states in simulation.

We probe local action responsiveness and generalization by varying actions, appearance, objects, and task requirements.
Controlled action comparisons isolate targeted arm or finger motions while keeping other parts stationary in the reference.
Appearance changes preserve actions and physical trajectories, while object replacements change the manipulated objects without changing task goals.
Novel tasks introduce new manipulation requirements using existing scenes and skills.

We evaluate multiple widely-used world models spanning general-purpose video generators, action-conditioned world models, and world action models.
The controlled tests reveal limitations in fine-grained action following.
Three of the four models evaluated for action control achieve at least 0.98 arm-side accuracy, yet none exceeds 0.30 on our finger-control tests. This gap motivates action representations that explicitly distinguish hands and individual fingers.
Predicted interaction outcomes also deviate from their references, with the best outcome consistency score reaching 0.737 (Table~\ref{tab:main_results}).
Generalization remains limited, with the highest mean object-preservation and novel-task-completion scores reaching only 2.52 and 2.02, respectively, on a five-point scale (Table~\ref{tab:generalization_results}). Qualitative comparisons further show that replacement objects can revert to familiar appearances after hand occlusion, motivating persistent modeling of object states through occlusion.

\noindent We summarize our main contributions below.
\begin{itemize}[leftmargin=*, labelsep=0.5em, topsep=3pt, itemsep=2pt, parsep=0pt]
    \item We introduce \textbf{\benchmarkname{}}, a diagnostic benchmark for fine-grained control and interaction fidelity in dexterous world models, complemented by evaluations of downstream execution utility and generalization.

    \item We structure the evaluation data to assess fine-grained action responsiveness, interaction outcome fidelity, and generalization.
    The evaluation sets comprise 220 full-task sequences, 90 critical interaction clips, 70 controlled action comparisons, and 380 generalization sequences.

    \item We evaluate widely-used world models under a common protocol, identifying limitations in fine-grained action following, interaction prediction, and generalization to novel objects and tasks to guide future work.
\end{itemize}

\section{Related Work}
\label{sec:related}

\subsection{World Models for Embodied Intelligence}
\label{sec:related_world_models_ei}
World models support embodied decision-making by predicting the dynamics of the environment. Latent-space approaches include DreamerV3~\citep{hafner2023mastering}, which optimizes policies through imagined trajectories, and V-JEPA 2~\citep{assran2025v}, which combines video representation pretraining with action-conditioned post-training for visually goal-conditioned robotic planning. Among video-generative approaches, UniSim~\citep{yang2023learning} is conditioned on instructions or low-level controls. Ctrl-World~\citep{guo2026ctrl} uses frame-wise action conditioning and multi-view prediction for policy evaluation and data generation, and DreamDojo~\citep{gao2026dreamdojo} transfers interaction priors from large-scale human videos to robotic action spaces for evaluation and planning. Building on these advances, we focus on dexterous manipulation and examine whether world models faithfully predict the interaction outcomes induced by fine-grained control inputs.

\subsection{World Models for Dexterous Manipulation}
\label{sec:related_dexterous_world_models}
Dexterous manipulation requires jointly modeling arm motion, finger configurations, and hand--object contact, as even subtle finger movements can determine interaction outcomes. DexWM~\citep{dexwm} represents actions using finger keypoints from human videos, predicts future latent states, and introduces hand-consistency constraints to improve finger-configuration modeling. DWM~\citep{kim2026dexterous} generates hand-action-driven interaction videos conditioned on 3D scene renderings and egocentric hand motion. DexAC-WM~\citep{yuan2026not} uses structured action conditioning to address scale disparities among components of high-dimensional actions and better capture subtle movements, while Mask2Real-WM~\citep{mask2realwm} decouples segmentation-mask prediction from appearance generation to learn robotic-hand action responses from simulation. Despite these advances, each method is evaluated under its own data and protocol, and there is still no shared benchmark that tests whether dexterous world models respond correctly to fine-grained hand commands and predict the resulting interaction outcomes.

\subsection{Benchmarks for Embodied World Models}
\label{sec:related_benchmarks}

Existing benchmarks address complementary aspects of embodied world models.
WorldArena~\citep{worldarena}, RoboWM-Bench~\citep{robowmbench}, and KineBench~\citep{kinebench} evaluate downstream utility or physical executability, with KineBench additionally testing generalization across appearance, objects, and tasks.
WMBench~\citep{gigaworld-1} assesses whether predicted outcomes agree with successful and failed real-world executions.
WorldSimProbe~\citep{co2026worldsimprobe} diagnoses action responsiveness and interaction grounding through controlled interventions, while H2R-Bench~\citep{h2rbench} evaluates functional interaction transfer across embodiments, including dexterous hands.

\benchmarkname{} evaluates whether world models follow action inputs for specific body parts, such as arms or fingers (Table~\ref{tab:benchmark_comparison}).
It also checks whether object trajectories and interaction outcomes match the reference, including failures.
Video-to-action execution and tests across appearance, objects, and tasks assess downstream utility and generalization.
In Table~\ref{tab:benchmark_comparison}, WorldArena, WorldSimProbe, and H2R-Bench partially cover outcome fidelity through aggregate success-rate agreement, interaction checks, and goal-preserving transfer, respectively.
H2R-Bench also partially covers generalization through embodiment transfer without a held-out-embodiment split.

\begin{table}[t]
\small
\centering
\setlength{\tabcolsep}{1.5pt}
\renewcommand{\arraystretch}{0.95}
\caption{
\textbf{Comparison of embodied world-model benchmarks.}
$\checkmark$: dedicated evaluation;
$\triangle$: partial or indirect coverage;
$\times$: no dedicated evaluation.
A/O/T = appearance/object/task.
}
\label{tab:benchmark_comparison}
\resizebox{\textwidth}{!}{
\begin{tabular}{@{}lcccccc@{}}
\toprule
\textbf{Benchmark}
& \textbf{\makecell[c]{Dexterous-Hand\\Evaluation}}
& \textbf{\makecell[c]{Local Action\\Probes}}
& \textbf{\makecell[c]{Object-Response\\Fidelity}}
& \textbf{\makecell[c]{Outcome Fidelity\\(Success \& Failure)}}
& \textbf{\makecell[c]{Video-to-Action\\Execution}}
& \textbf{\makecell[c]{Generalization\\Tests}} \\
\midrule

WorldArena~\citep{worldarena}
& $\times$ & $\times$ & $\triangle$ & $\triangle$ & $\checkmark$ & $\times$ \\

WMBench~\citep{gigaworld-1}
& $\times$ & $\times$ & $\triangle$ & $\checkmark$ & $\times$ & $\times$ \\

RoboWM-Bench~\citep{robowmbench}
& $\times$ & $\times$ & $\times$ & $\times$ & $\checkmark$ & $\times$ \\

KineBench~\citep{kinebench}
& $\times$ & $\times$ & $\times$ & $\times$ & $\checkmark$
& $\checkmark$\, A/O/T \\

WorldSimProbe~\citep{co2026worldsimprobe}
& $\times$ & $\checkmark$\, Arm perturbation
& $\checkmark$ & $\triangle$ & $\times$
& $\checkmark$\, Action/scene \\

H2R-Bench~\citep{h2rbench}
& $\checkmark$ & $\times$ & $\checkmark$ & $\triangle$ & $\times$
& $\triangle$\, Embodiment \\

\midrule
\textbf{\benchmarkname{} (Ours)}
& $\checkmark$ & $\checkmark$\, Arm/finger isolation & $\checkmark$
& $\checkmark$ & $\checkmark$
& $\checkmark$\, A/O/T \\

\bottomrule
\end{tabular}
}
\end{table}

\section{The \benchmarkname{} Benchmark}
\label{sec:benchmark}

\subsection{Benchmark Design}
\label{sec:benchmark_design}

\benchmarkname{} evaluates dexterous world models in terms of \emph{fine-grained control accuracy}, \emph{interaction fidelity}, and \emph{downstream execution utility}. It uses three core evaluation sets, namely controlled action comparisons, critical interaction clips, and full task sequences, built from the original DexJoCo dataset~\citep{dexjoco} and from Cosmos 3 and Fast-WAM policy rollouts~\citep{cosmos3, fastwam}.

\paragraph{Controlled action comparisons.}
To test whether generated videos realize a command only in the targeted body part, we construct arm-level and finger-level comparisons in which all non-target parts are held still.
For the arm level, we build 50 comparisons (25 left--right pairs) from five episodes of each of five bimanual tasks, namely Assembly, Hanoi, Microwave Cook, Photograph, and Unlock iPad, by action-space editing.
The two clips of a pair share an identical prefix. After a short interpolated bridge, one replays the recorded right-arm segment and the other the left-arm segment, with the opposite arm frozen at its initial pose.
The edited actions are re-rendered in the simulator, so that the clips of a pair differ only in the acting arm.
For the finger level, starting from frames of Pinch Tongs and Photograph in which all fingers are clearly visible, we fully extend one finger at a time while keeping the arm and the remaining fingers stationary.

\paragraph{Critical interaction clips.}
To evaluate how faithfully world models predict critical hand-object interactions and their outcomes, we construct a short-clip evaluation set covering grasping, insertion, tool use and bimanual manipulation.
We select successful samples from the original DexJoCo dataset and failed samples from policy rollouts.
We then manually identify the start and end points of each clip, extracting approximately 1-14 seconds (median 3.6 s) of the critical interaction.
Together, successful and failed samples test whether generated videos preserve the interaction dynamics and outcomes of their corresponding reference sequences.
Faithfully reproducing a failed interaction therefore also demonstrates predictive fidelity.

\paragraph{Full task sequences.}
For each of the 11 tasks, we use episodes 0--89 for fine-tuning and hold out episodes 90--99 for evaluation.
We evaluate 10 training-set and 10 held-out full task sequences per task, yielding 110 per subset.

\paragraph{Generalization samples.}
We construct three additional evaluation sets for appearance, object, and task generalization.
\emph{Appearance generalization.} We select 110 sequences from the original dataset and re-render each under new texture and lighting conditions separately, while keeping the actions, physical states, and camera settings unchanged. This yields 220 sequences with appearance variations.
\emph{Object generalization.} We introduce five object replacement conditions in existing tasks, covering variations in shape, appearance, and object category. These conditions use Sprite bottles, tennis balls, eyeglasses, pegs and sockets, and rings for the Tower of Hanoi task. For each condition, we adapt the grasping and manipulation actions of 20 original demonstrations and re-simulate them, yielding 100 trajectories in total.
\emph{Task generalization.} Using existing scenes and manipulation skills, we construct three new tasks involving object retrieval from a bucket, bimanual disassembly, and object retrieval from a microwave. For each task, we vary the initial object positions, orientations, and manipulation endpoints to generate 20 simulated trajectories, yielding 60 trajectories in total.

\subsection{Fine-Grained Control Accuracy}
\label{sec:benchmark_control}

Arm-side and finger control accuracy apply only to models that take the tested robot actions as input; models without such inputs are marked N/A for these two metrics.

\paragraph{Hand Geometry Accuracy.}
Hand geometry accuracy measures whether the dexterous hand in a generated video can be explained by a kinematically valid Allegro hand.
A YOLOv8-pose detector~\citep{ultralyticsyolov8}, fine-tuned on simulator-rendered frames, localizes 20 keypoints per hand. A pose fitter then optimizes the 16 joint angles and the palm pose so that the forward kinematics of the simulator's Allegro model (MJCF) reprojects onto the detected 2D keypoints.
We compute four case-level measurements: detection rate, reprojection residual, joint-limit saturation, and self-collision depth.
Detection contributes its observed rate; the three errors are mapped to bounded scores using fixed ground-truth training-set 95th-percentile scales, with capped reciprocals for reprojection and saturation and a smooth reciprocal for collision.
Cases with detection rates below 0.8 or no scored observations contribute zero to the three geometric components.
Each component is averaged over its fixed eligible case set, followed by equal averaging across the four components and the training-set and held-out subsets.

\paragraph{Arm-Side Control Accuracy.}
Arm-side control accuracy measures whether a generated video exhibits the intended motion only on the side specified by the instruction.
The judgment is made by an MLLM judge~\citep{llm-as-judge, mllm-as-judge}.
To avoid leaking the answer, the judge observes only the video without being told the motion instruction. It separately determines on which side of the view the manipulation occurs, which of the robot's own hands performs it, and which disk size is moved.
Each question is answered over multiple samples with a majority vote, and the votes are compared against the instructed target side to obtain the accuracy.

\paragraph{Finger Control Accuracy.}
Finger control accuracy measures whether a generated video produces the intended motion only in the instructed finger, while the remaining fingers and the arm stay still.
An MLLM judge observes the video without being told that any finger will move, answers which finger extends with a joint hand--finger label, and the judgment is made by a majority vote over multiple samples.
This joint label captures both the response of the target finger and the isolation of the remaining body parts.
The judgment is compared against ground-truth labels recorded during construction to obtain the accuracy, requiring no manual annotation.
Unperturbed original clips, for which no finger extension is commanded, serve as negative controls to test whether the judge fabricates non-existent motion.

\subsection{Interaction Fidelity}
\label{sec:benchmark_interaction}

\paragraph{Object Trajectory Score.}
The Object Trajectory Score (OTS) measures the agreement between object motion in generated videos and reference trajectories.
We establish and verify two-dimensional reference trajectories on the original ground-truth videos and track the corresponding initial points in generated videos using CoTracker3~\citep{cotracker3}.
Tracking and evaluation for each point are restricted to its predefined visible prefix.
Valid observations receive a bounded reciprocal score based on positional error and a GT-calibrated scale.
Missing observations contribute zero, while the denominator remains fixed at the number of reference-defined eligible point--frame pairs.
Scores are averaged over objects within each case, cases within each task, and tasks within each split, followed by an equal average over the ID and OOD splits.

\paragraph{Interaction Outcome Consistency.}
Interaction outcome consistency measures how consistently the interaction outcome presented in a generated video matches that of the reference sequence.
An MLLM judge rates the degree of interaction execution in the video on a 1--5 scale against task-specific anchors that describe observable interaction outcomes in the reference clips.
The absolute difference between the scores of a generated video and its reference sequence gives the fidelity error of the sample, which is mapped to a per-sample score of $1 - |s_{\mathrm{gen}} - s_{\mathrm{ref}}|/4$, so that 1 indicates a matching outcome. Successful and failed samples are compared under the same criterion, and scores are averaged over all samples.

\subsection{Downstream Execution Utility}
\label{sec:benchmark_execution}

We evaluate the utility of generated videos for downstream task execution.
Following prior work~\citep{unipi,lvp,vidar}, we compose a video generator $G$ with an action decoder $I$ to form an execution policy, $\pi = I \circ G$.
The generator produces a visual trajectory under the specified generation conditions, and the decoder maps it to robot controls.
We implement the decoder using an inverse dynamics model and keep it fixed across video models within each task.
This dimension evaluates the resulting policy through simulation execution and agreement with reference actions.
Appendix~\ref{sec:app_idm} details decoder setup, calibration, and sources of execution error.

\paragraph{Execution Success Rate.}
Execution Success Rate evaluates whether imagined video plans support successful task completion.
Given a task, the video model imagines its execution as a visual trajectory, and an action decoder translates this plan into robot controls.
We evaluate task completion from selected intermediate states by executing the corresponding controls open-loop in simulation.
We report the proportion of trials satisfying predefined physical success criteria, averaged equally across five single-arm manipulation tasks and two intermediate-state initialization settings.

\paragraph{Action Recovery Fidelity.}
Action Recovery Fidelity measures how closely the controls recovered from generated videos match the recorded reference actions.
We temporally align the decoded and reference sequences and assess end-effector position, arm orientation, and finger joint targets separately.
Ground-truth videos processed by the same action decoder provide calibration references for the aggregate score.
This metric complements task-level success by assessing recovered control sequences in detail.

\subsection{Auxiliary Visual Metrics}
Beyond the three core dimensions, five conventional visual quality metrics are included as auxiliary references, each computed frame by frame and averaged over the sequence.
PSNR and SSIM~\citep{ssim} measure the pixel-level and structural similarity between generated and reference frames, and LPIPS~\citep{lpips} measures the perceptual distance with AlexNet features. All three are computed after frame-by-frame alignment with the reference video, excluding the first frame that serves as the generation condition.
The imaging quality score, based on the MUSIQ~\citep{musiq} model, provides a no-reference assessment of per-frame imaging quality.
In addition, the cosine similarity of DINO \citep{dino} features to the first frame and to the previous frames measures the temporal consistency of video appearance, conflating long-term drift and short-term jitter into a single score.

\subsection{Generalization Evaluation}
\label{sec:benchmark_generalization}

Generalization evaluation measures the ability of world models to preserve the specified conditions, complete new tasks, and maintain visual integrity under appearance, object, and task variations.
We use a multimodal large language model (MLLM) to evaluate the generated videos.
Appearance generalization includes texture and lighting variations, assessing whether the specified textures and lighting are preserved in subsequent frames.
Object generalization checks whether replacement objects retain their identities, shapes, and key attributes.
Task generalization assesses the completion of key task stages and the final goal.
The judge assigns a score from 1 to 5 for each generalization category using the initial input frame, subsequent sampled frames, and the corresponding assessment criteria.
Across all categories, it also checks for visual collapse, including subject disappearance unexplained by normal occlusion, severe structural fragmentation, and abrupt scene changes.
We report the mean generalization score and the fraction of videos with visual collapse in each category.

\begin{table}[t]
\centering
\caption{Main results on \benchmarkname{} across fine-grained control accuracy,
interaction fidelity, and downstream execution utility.
$\uparrow$ indicates higher is better.
Best results are in \textbf{bold}.
N/A: Wan2.2-TI2V and FastWAM do not take robot actions as input, so the arm-side and finger-level control evaluations do not apply.}
\label{tab:main_results}

\begingroup
\small
\setlength{\tabcolsep}{2pt}
\renewcommand{\arraystretch}{1.15}
\begin{tabular*}{\textwidth}{@{\extracolsep{\fill}}lccccccc@{}}
\toprule
\textbf{Model}
& \multicolumn{3}{c}{\textbf{Fine-Grained Control Accuracy}}
& \multicolumn{2}{c}{\textbf{Interaction Fidelity}}
& \multicolumn{2}{c}{\textbf{Downstream Execution Utility}} \\
\cmidrule(lr){2-4}
\cmidrule(lr){5-6}
\cmidrule(lr){7-8}

& \shortstack{Hand Geom.\\Accuracy $\uparrow$}
& \shortstack{Arm-Side Ctrl.\\Accuracy $\uparrow$}
& \shortstack{Finger Ctrl.\\Accuracy $\uparrow$}
& \shortstack{Object Traj.\\Score $\uparrow$}
& \shortstack{Outcome\\Consistency $\uparrow$}
& \shortstack{Execution\\Success Rate $\uparrow$}
& \shortstack{Action Recovery\\Fidelity $\uparrow$} \\
\midrule

Wan2.2-TI2V \citep{wan} & 0.253 & N/A & N/A & 0.220 & 0.638 & 0.140 & 0.087 \\
IRASim  \citep{irasim}    & 0.570 & 0.98 & \textbf{0.30} & 0.458 & \textbf{0.737} & \textbf{0.615} & 0.326 \\
Ctrl-World \citep{guo2026ctrl} & \textbf{0.808} & 0.98 & 0.10 & 0.488 & 0.716 & 0.590 & 0.320 \\
FastWAM \citep{fastwam}    & 0.802 & N/A & N/A & \textbf{0.534} & 0.716 & 0.315 & 0.109 \\
Cosmos3 \citep{cosmos3}    & 0.806 & 0.48 & 0.10 & 0.475 & 0.674 & 0.410 & 0.131 \\

Mask2Real-WM \citep{mask2realwm} & 0.511 & \textbf{1.00} & 0.05 & 0.461 & 0.646 & 0.575 & \textbf{0.351} \\
\bottomrule
\end{tabular*}
\endgroup
\end{table}

\section{Experiments}
\label{sec:experiments}

\subsection{Experimental Setup}
\label{sec:exp_setup}

\paragraph{Datasets.}
Our primary evaluation uses the DexJoCo-based~\citep{dexjoco} evaluation sets of \benchmarkname{} (Section~\ref{sec:benchmark_design}).
We additionally report real-robot evaluations in Section~\ref{sec:exp_real_robot}.

\paragraph{Baselines.}
We evaluate six representative models spanning three categories: a general-purpose video generator (Wan2.2-TI2V~\citep{wan}), three action-conditioned world models (IRASim~\citep{irasim}, Ctrl-World~\citep{guo2026ctrl}, and Mask2Real-WM~\citep{mask2realwm}), and two world action models (FastWAM~\citep{fastwam} and Cosmos3~\citep{cosmos3}).
Among these baselines, Mask2Real-WM is specifically designed for dexterous robotic manipulation.
For the core DexJoCo evaluation, all baselines are fine-tuned on the same training data.
Wan2.2-TI2V and FastWAM do not take robot actions as input in our evaluation and are therefore excluded from the arm-side and finger-level control evaluations.
All methods use the same frame rate and number of camera viewpoints in the core evaluation.
Training and generation details are provided in Appendix~\ref{sec:app_baseline_config}.

\subsection{Main Results}
\label{sec:exp_main}

\subsubsection{Overall Performance on \benchmarkname{}}
Table~\ref{tab:main_results} reveals two main findings. First, action-conditioned models outperform the general-purpose video generation baseline in interaction prediction and downstream execution utility, while exhibiting different strengths across metrics. IRASim achieves the highest Outcome Consistency (0.737) and Execution Success Rate (0.615), whereas Mask2Real-WM leads in Action Recovery Fidelity (0.351). Across all six baselines, FastWAM obtains the highest Object Trajectory Score (0.534). These results highlight the complementary information provided by trajectory agreement, interaction outcomes, and downstream execution utility.

Second, action conditioning alone does not guarantee fine-grained action control. Mask2Real-WM, IRASim, and Ctrl-World achieve arm-side control accuracies of 1.00, 0.98, and 0.98, respectively. Cosmos3 scores only 0.48 and, in the qualitative comparison, reproduces the training-time motion pattern rather than following the commanded arm. Finger-level control remains challenging for all four models, with accuracies at or below 0.30. In particular, Mask2Real-WM achieves perfect arm-side accuracy on the evaluated set but only 0.05 in finger control. This gap motivates grouping joint commands within each finger and explicitly encoding hand and finger identities to better align each finger's actions with its corresponding visual region.

\subsubsection{Fine-Grained Control Accuracy}
The results reveal distinct levels of performance in hand geometry, arm-side responses, and isolated finger control, with the last remaining challenging across models.

\begin{figure}[t]
    \centering
    \begingroup
    \setlength{\tabcolsep}{1pt}
    \begin{tabular}{@{}ccccccc@{}}
\includegraphics[width=.137\textwidth]{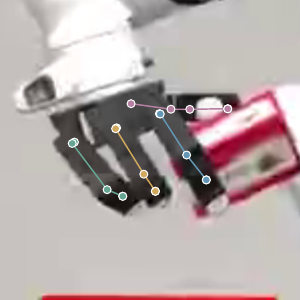} &
\includegraphics[width=.137\textwidth]{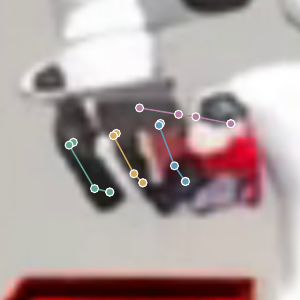} &
\includegraphics[width=.137\textwidth]{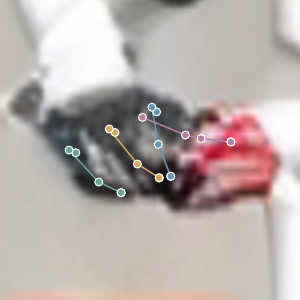} &
\includegraphics[width=.137\textwidth]{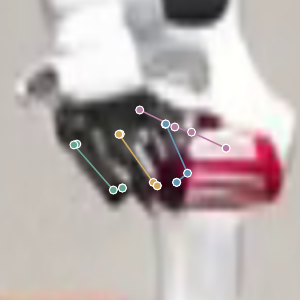} &
\includegraphics[width=.137\textwidth]{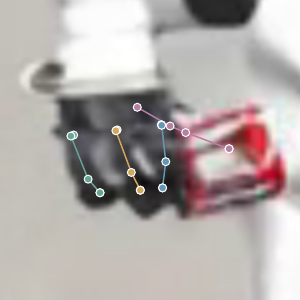} &
\includegraphics[width=.137\textwidth]{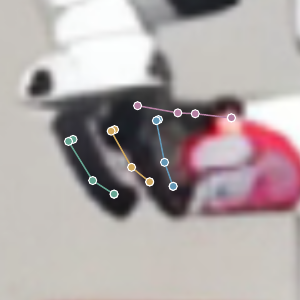} &
\includegraphics[width=.137\textwidth]{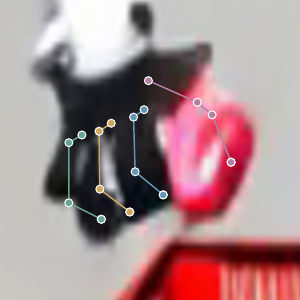} \\[2pt]
\includegraphics[width=.137\textwidth]{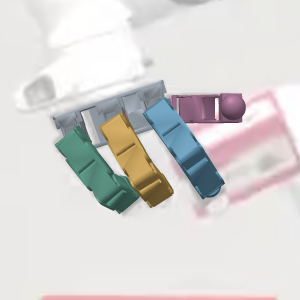} &
\includegraphics[width=.137\textwidth]{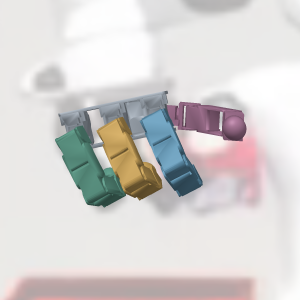} &
\includegraphics[width=.137\textwidth]{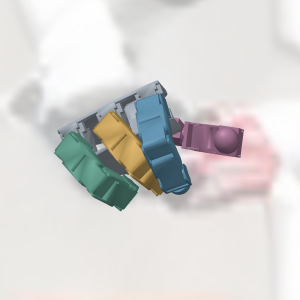} &
\includegraphics[width=.137\textwidth]{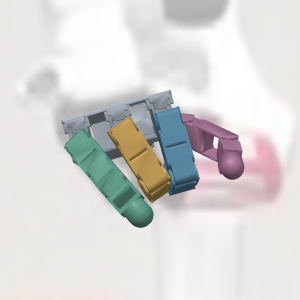} &
\includegraphics[width=.137\textwidth]{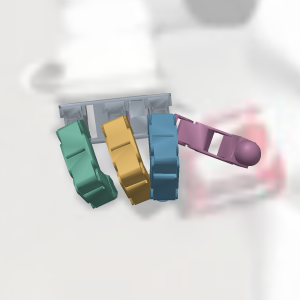} &
\includegraphics[width=.137\textwidth]{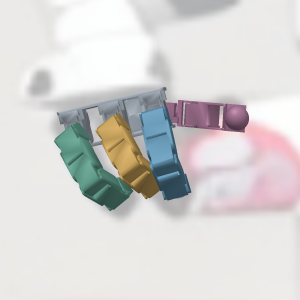} &
\includegraphics[width=.137\textwidth]{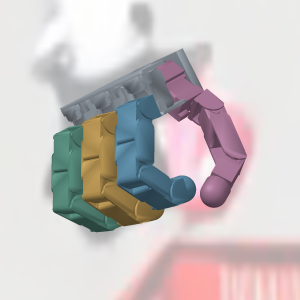} \\[2pt]
{\scriptsize GT} & {\scriptsize Ctrl-World} & {\scriptsize Cosmos3} & {\scriptsize FastWAM} & {\scriptsize IRASim} & {\scriptsize Mask2Real-WM} & {\scriptsize Wan2.2} \\
    \end{tabular}
    \endgroup
    \caption{Qualitative comparison of hand geometry accuracy. Columns show GT and six baselines. The top row shows video crops with detected 2D hand keypoints. The bottom row overlays the fitted 3D hand on the same crops. Colours distinguish fingers.}
    \label{fig:hand_pose_detail}
\end{figure}

\paragraph{Hand Geometry Accuracy.}
Ctrl-World, Cosmos3, and FastWAM obtain closely matched HGA scores of 0.808, 0.806, and 0.802, respectively, followed by IRASim at 0.570, Mask2Real-WM at 0.511, and Wan2.2-TI2V at 0.253. These scores jointly reflect hand detection coverage and estimated geometric quality. The strong HGA of Cosmos3 contrasts with its limited arm-side responsiveness, illustrating that geometric plausibility alone does not establish adherence to action commands. Fig.~\ref{fig:hand_pose_detail} shows the generated hand configurations and their fitted poses.

\begin{figure}[t]
    \centering
    \captionsetup{skip=4pt}
    \begingroup
    \footnotesize
    \setlength{\tabcolsep}{2pt}
    \begin{tabular}{@{}ccccc@{}}
\includegraphics[width=.19\textwidth]{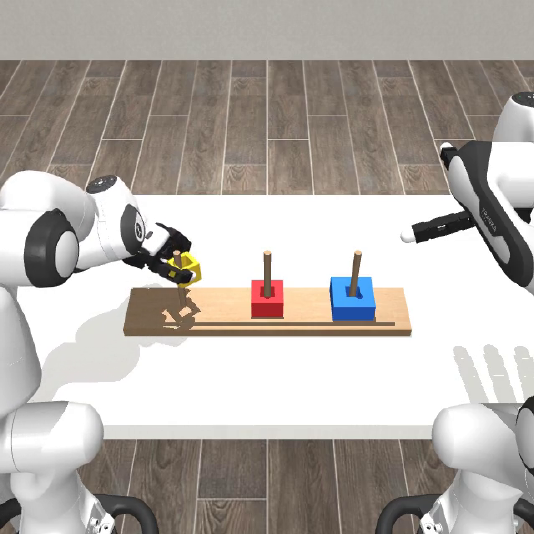} &
\includegraphics[width=.19\textwidth]{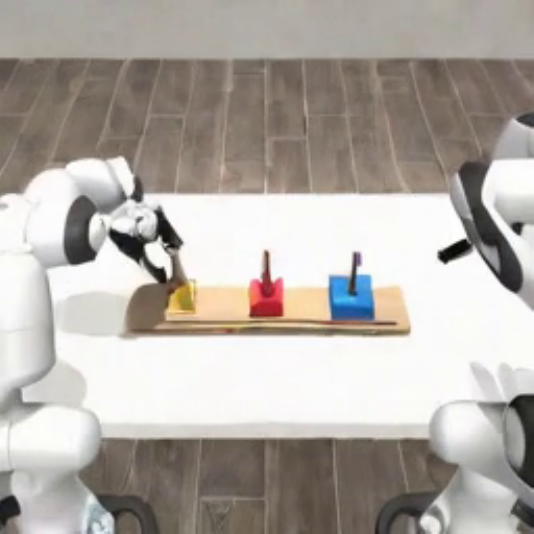} &
\includegraphics[width=.19\textwidth]{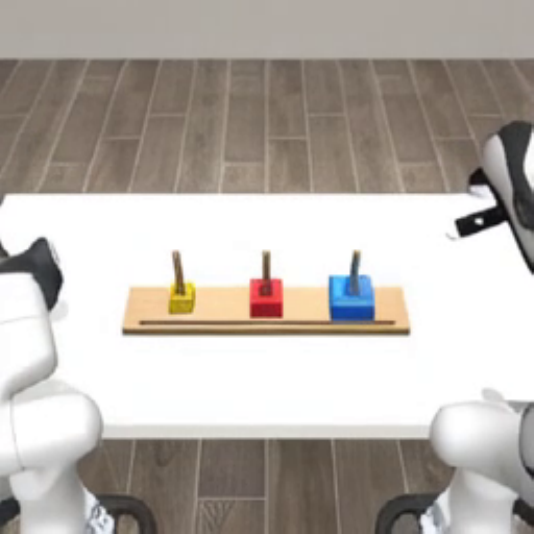} &
\includegraphics[width=.19\textwidth]{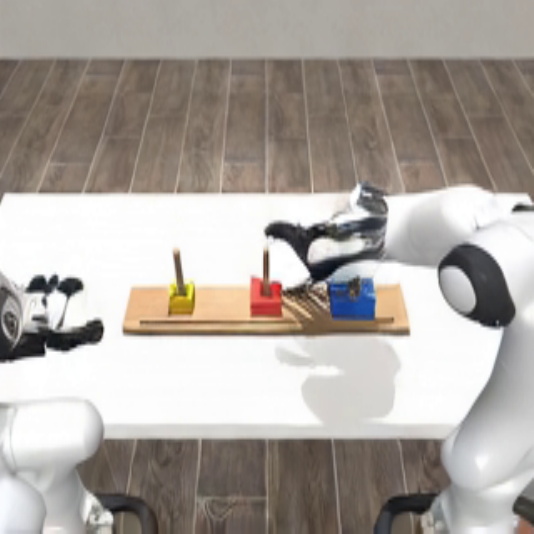} &
\includegraphics[width=.19\textwidth]{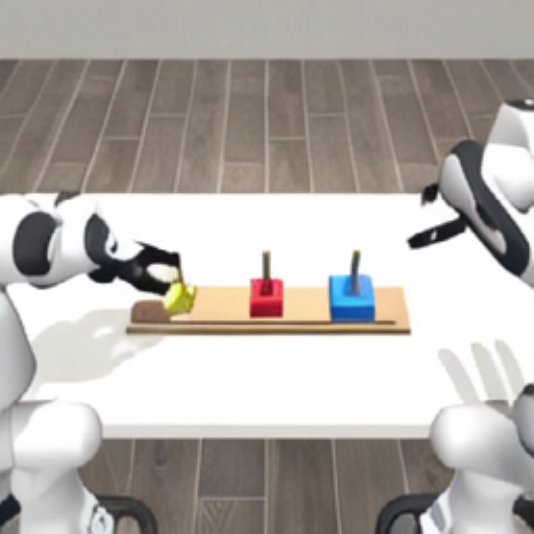} \\
GT & IRASim & Ctrl-World & Cosmos3 & Mask2Real-WM \\
    \end{tabular}
    \endgroup
    \caption{Qualitative comparison of the final frames generated by different models in the arm-side control accuracy evaluation, given an action command specifying left-arm motion. The leftmost panel shows the ground-truth frame.}
    \label{fig:arm_side_left_8s}
\end{figure}

\paragraph{Arm-Side Control Accuracy.}
With balanced left- and right-arm commands, always selecting the same arm yields an accuracy of \(0.5\). Mask2Real-WM reaches \(1.00\), while IRASim and Ctrl-World each achieve \(0.98\), indicating reliable arm-side responses on this evaluation set. Cosmos3 scores only \(0.48\). The qualitative comparison in Fig.~\ref{fig:arm_side_left_8s} illustrates this contrast: Mask2Real-WM, IRASim, and Ctrl-World move the commanded left arm while keeping the right arm still, whereas Cosmos3 moves the right arm, which always moves first in the Hanoi training demonstrations. Its behavior here reproduces the training-time motion pattern rather than following the commanded arm.

\paragraph{Finger Control Accuracy.}
All four evaluated models respond poorly to individual-finger commands. IRASim achieves the highest accuracy of \(0.30\), followed by Ctrl-World and Cosmos3 at \(0.10\) and Mask2Real-WM at \(0.05\). The generated hands often reproduce coordinated whole-hand motions rather than independently moving the specified finger. These results identify isolated finger control as a persistent limitation, even for models that reliably respond to arm-side commands.

\subsubsection{Interaction Fidelity}

\begin{figure}[t]
\centering
\begingroup
\setlength{\tabcolsep}{1pt}
\fontsize{8}{10}\selectfont
\begin{tabular}{@{}*{7}{>{\centering\arraybackslash}m{0.138\textwidth}}@{}}
\includegraphics[width=\linewidth]{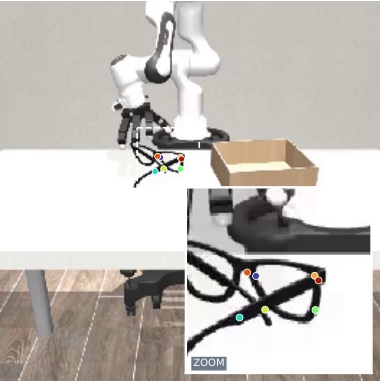} & \includegraphics[width=\linewidth]{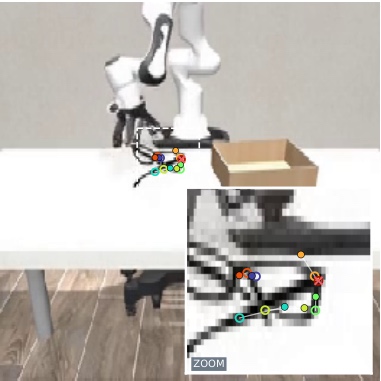} & \includegraphics[width=\linewidth]{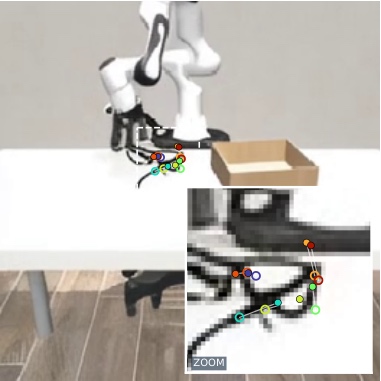} & \includegraphics[width=\linewidth]{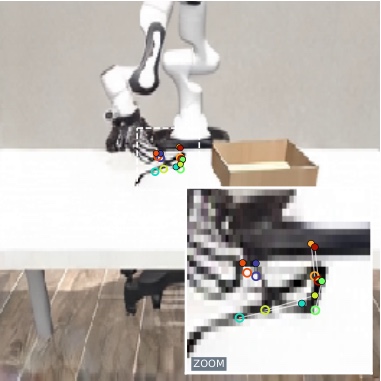} & \includegraphics[width=\linewidth]{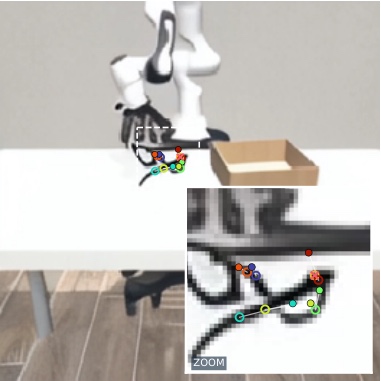} & \includegraphics[width=\linewidth]{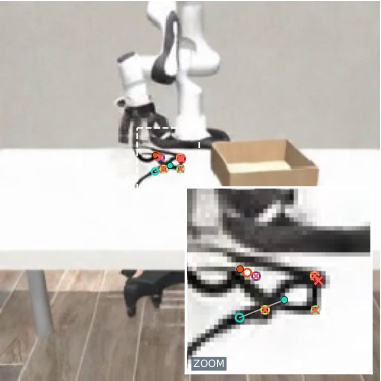} & \includegraphics[width=\linewidth]{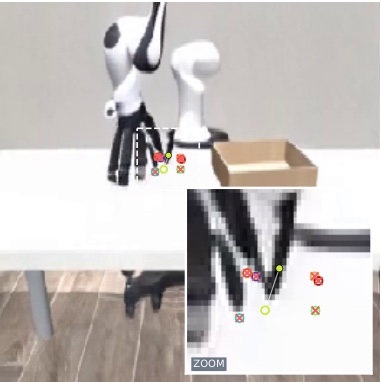} \\[3pt]
\shortstack{\textbf{GT}\\Reference} & \shortstack{\textbf{FastWAM}\\OTS: 0.5008} & \shortstack{\textbf{Ctrl-World}\\OTS: 0.5085} & \shortstack{\textbf{Cosmos3}\\OTS: 0.4351} & \shortstack{\textbf{Mask2Real-WM}\\OTS: 0.4797} & \shortstack{\textbf{IRASim}\\OTS: 0.4464} & \shortstack{\textbf{Wan}\\OTS: 0.4063} \\
\end{tabular}
\endgroup
\caption{Qualitative illustration of the Object Trajectory Score (OTS). GT and six baselines are shown at 6.6\,s for a single \emph{fold glasses} case, with tracked-point overlays and zoomed-in views. The scores below the model names are the OTS values for this case.}
\label{fig:object_traj_fold_glasses}
\end{figure}

\paragraph{Object Trajectory Consistency.}
We evaluate object motion using the Object Trajectory Score over reference-defined visible intervals. Under the current reference protocol, FastWAM achieves the highest score of \(0.534\), followed by Ctrl-World at \(0.488\) and Cosmos3 at \(0.475\). Mask2Real-WM and IRASim obtain closely matched scores of \(0.461\) and \(0.458\), while Wan2.2-TI2V scores \(0.220\). Fig.~\ref{fig:object_traj_fold_glasses} illustrates differences in tracked object motion for a fold-glasses case; its case-level scores need not follow the aggregate model ranking.
Additional timestamps are shown in Appendix~\ref{sec:app_ots_visualization}.

\paragraph{Outcome Consistency.}
IRASim achieves the highest Outcome Consistency of \(0.737\), followed by Ctrl-World and FastWAM tied at \(0.716\), Cosmos3 at \(0.674\), Mask2Real-WM at \(0.646\), and Wan2.2-TI2V at \(0.638\). The different rankings between outcome consistency and trajectory metrics indicate that accurate local object motion does not necessarily preserve the final interaction outcome.

\begin{table}[b]
\centering
\caption{Auxiliary visual metrics of the six baselines. The best model in each column is in \textbf{bold}.}
\label{tab:visual_quality}
\begingroup
\small
\setlength{\tabcolsep}{8pt}
\renewcommand{\arraystretch}{1.2}
\begin{tabular}{lccccc}
\toprule
\multicolumn{1}{c}{\multirow{2}{*}[-4pt]{\textbf{Model}}}
& \multicolumn{2}{c}{\textbf{Reference-Free}}
& \multicolumn{3}{c}{\textbf{Paired with GT Frame}} \\
\cmidrule(lr){2-3}
\cmidrule(lr){4-6}

& \makecell[c]{Imaging\\Quality $\uparrow$}
& \makecell[c]{Subject\\Consistency $\uparrow$}
& PSNR $\uparrow$
& SSIM $\uparrow$
& LPIPS $\downarrow$ \\
\midrule

Wan2.2-TI2V \citep{wan} & 0.426 & 0.881 & 13.98 & 0.662 & 0.442 \\
IRASim \citep{irasim}    & 0.453 & 0.921 & 20.09 & 0.788 & 0.248 \\
Ctrl-World \citep{guo2026ctrl} & 0.495 & \textbf{0.930} & 19.02 & 0.781 & 0.251 \\
FastWAM \citep{fastwam}    & 0.524 & 0.929 & \textbf{21.30} & \textbf{0.826} & \textbf{0.200} \\
Cosmos3 \citep{cosmos3}    & \textbf{0.538} & 0.918 & 18.45 & 0.781 & 0.239 \\
Mask2Real-WM \citep{mask2realwm} & 0.479 & 0.919 & 20.95 & 0.794 & 0.241 \\
\midrule
Real video (reference) & 0.533 & 0.927 & -- & -- & -- \\

\bottomrule
\end{tabular}
\endgroup
\end{table}

\subsubsection{Downstream Execution Utility}
Table~\ref{tab:main_results} reports Execution Success Rate and Action Recovery Fidelity for each video model paired with the action decoder. Averaged over the intermediate-state initialization settings and five single-arm tasks, IRASim achieves the highest Execution Success Rate of \(0.615\), followed by Ctrl-World at \(0.590\) and Mask2Real-WM at \(0.575\). Cosmos3, FastWAM, and Wan2.2-TI2V obtain \(0.410\), \(0.315\), and \(0.140\), respectively. Mask2Real-WM achieves the highest Action Recovery Fidelity of \(0.351\), followed by IRASim at \(0.326\). The two metrics provide complementary evidence: Action Recovery Fidelity measures agreement with reference controls, whereas Execution Success Rate measures whether the imagined video plan, translated into controls by the action decoder, supports task completion through open-loop simulation execution.

\subsubsection{Visual quality and manipulation fidelity}

Table~\ref{tab:visual_quality} reports auxiliary visual metrics, averaged across the training-set and held-out test-set results. FastWAM leads all three GT-paired metrics, achieving a PSNR of 21.30, an SSIM of 0.826, and an LPIPS of 0.200. Cosmos3 scores highest on imaging quality. Ctrl-World and FastWAM show the strongest subject consistency, with scores of 0.930 and 0.929, respectively, close to the real-video score of 0.927. Wan2.2-TI2V ranks last across all five metrics.

\subsubsection{Generalization}
\label{sec:exp_generalization}
We evaluate appearance and object preservation, new task completion, and visual failures using the MLLM judge described in Section~\ref{sec:benchmark_generalization}, with results reported in Table~\ref{tab:generalization_results}.
Qualitative comparisons are provided in Appendix~\ref{sec:app_generalization}.

\begin{table}[b]
\centering
\caption{Generalization results across texture, lighting, object, and task variations. Each cell reports the generalization score ($\uparrow$) / visual collapse rate ($\downarrow$), with scores measuring texture and lighting preservation, object preservation, or task completion, respectively. Bold and underlined values indicate the best and second-best results.}
\label{tab:generalization_results}
\begingroup
\small
\setlength{\tabcolsep}{6pt}
\renewcommand{\arraystretch}{1.15}
\begin{tabular*}{\textwidth}{@{\extracolsep{\fill}}lcccc@{}}
\toprule
\textbf{Model}
& \textbf{Texture}
& \textbf{Lighting}
& \textbf{Object}
& \textbf{Task} \\
\midrule
Wan2.2-TI2V \citep{wan} & \textbf{3.973} / 16.4\% & \underline{4.918} / 18.2\% & \textbf{2.520} / 73.0\% & 1.917 / 81.7\% \\
IRASim \citep{irasim} & 2.273 / 50.0\% & 4.618 / 25.5\% & 1.900 / 84.0\% & \underline{2.000} / 100.0\% \\
Ctrl-World \citep{guo2026ctrl} & 2.573 / \textbf{0.9\%} & 4.655 / \textbf{3.6\%} & \underline{2.350} / \textbf{33.0\%} & 1.883 / \textbf{63.3\%} \\
FastWAM \citep{fastwam} & 1.355 / 80.0\% & 2.945 / 32.7\% & 2.250 / \underline{38.0\%} & \textbf{2.017} / 73.3\% \\
Cosmos3 \citep{cosmos3} & 1.727 / 96.4\% & 3.227 / 86.4\% & 2.050 / 100.0\% & 1.967 / 98.3\% \\
Mask2Real-WM \citep{mask2realwm} & \underline{2.836} / \underline{2.7\%} & \textbf{4.991} / \underline{5.5\%} & 2.140 / 62.0\% & \textbf{2.017} / \underline{71.7\%} \\
\bottomrule
\end{tabular*}
\endgroup
\end{table}

\paragraph{Appearance generalization.}
Table~\ref{tab:generalization_results} shows that all models preserve lighting better than texture.
Wan achieves the highest texture score (3.973) and a lighting score of 4.918, suggesting that initial-frame conditioning and pretrained visual priors can support appearance preservation even without action inputs.
Ctrl-World achieves the lowest collapse rates for texture and lighting (0.9\% and 3.6\%), potentially benefiting from frame-level action conditioning and visual history.
Mask2Real-WM achieves the highest lighting score (4.991) and the second-highest texture score (2.836), with collapse rates of 5.5\% and 2.7\%, respectively.
Mask2Real-WM's low collapse rates under appearance changes suggest that mask constraints can help separate robot motion from scene appearance, stabilizing arm and hand prediction.

\paragraph{Object and task generalization.}
Transfer remains limited under both variations: object preservation scores range from 1.900 to 2.520, while task completion scores range from 1.883 to 2.017, with task collapse rates reaching 63.3\%--100.0\%.
Objects reverting to familiar appearances after occlusion motivate preserving object identity across frames and explicitly maintaining object states and contact with the hand during occlusion.
Even Ctrl-World, which achieves the lowest object and task collapse rates, scores only 2.350 and 1.883, respectively, further indicating that visual stability in generated videos alone does not ensure object preservation or successful completion of a novel task.

\subsection{Real-Robot Evaluation of Downstream Execution Utility}
\label{sec:exp_real_robot}

\begin{figure}[t]
\centering
\begin{minipage}[t]{0.35\linewidth}
  \centering
  \includegraphics[width=\linewidth]{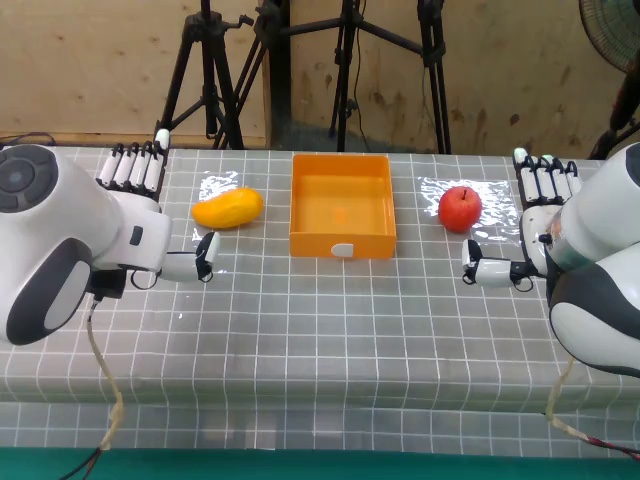}\\[2pt]
  \small Place Fruit into Box
\end{minipage}\quad
\begin{minipage}[t]{0.35\linewidth}
  \centering
  \includegraphics[width=\linewidth]{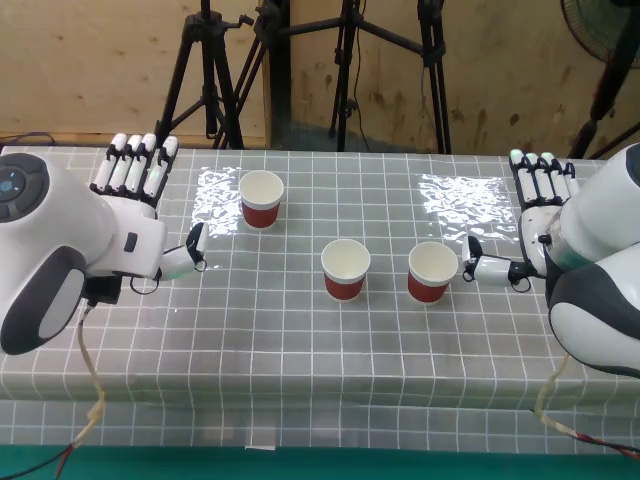}\\[2pt]
  \small Stack Cups
\end{minipage}
\caption{Initial egocentric frames for the two real-world tasks.}
\label{fig:real_robot_tasks}
\end{figure}

We evaluate two real-world tabletop tasks on a dual-arm FR3 platform equipped with XHand dexterous hands (Fig.~\ref{fig:real_robot_tasks}).
In \textit{Place Fruit into Box}, the robot places an apple and a mango in a box, and success requires both fruits to remain inside after release.
In \textit{Stack Cups}, the robot nests three cups, and success requires all three to remain stably nested after release.

Following the downstream execution protocol in Sec.~\ref{sec:benchmark_execution}, we use a fixed inverse dynamics model (IDM) for each task to decode actions from recorded demonstrations and videos generated by Cosmos3, Ctrl-World, and IRASim.
We execute the decoded actions open-loop on the robot and score each trial using the task criteria above (Table~\ref{tab:real_robot_idm_execution}).

\begingroup
\setlength{\intextsep}{6pt plus 2pt minus 2pt}
\begin{table}[!h]
\centering
\small
\caption{Real-robot execution success by task and video source, reported as successes out of ten trials.}
\label{tab:real_robot_idm_execution}
\begin{tabular}{lcccc}
\toprule
\textbf{Task} & \textbf{Demonstration} & \textbf{Cosmos3} & \textbf{Ctrl-World} & \textbf{IRASim} \\
\midrule
Place Fruit into Box & 10/10 & 5/10 & 4/10 & 3/10 \\
Stack Cups           & 9/10  & 6/10 & 7/10 & 4/10 \\
\bottomrule
\end{tabular}
\end{table}

In \textit{Place Fruit into Box} (Fig.~\ref{fig:real_robot_rollout_case_study}), Cosmos3 and Ctrl-World miss the finger closure seen in GT, and their IDM replays lift the hand before securing the mango.
The demonstration-driven replay places the mango with the same IDM.
IRASim distorts the mango at 9.0\,s and shows an empty box at 17.2\,s.
In \textit{Stack Cups}, at 18.0\,s Ctrl-World loses the upper cup's red coloring after release and IRASim distorts the left cup's rim.

\begin{figure}[!h]
\centering
\includegraphics[width=0.96\linewidth]{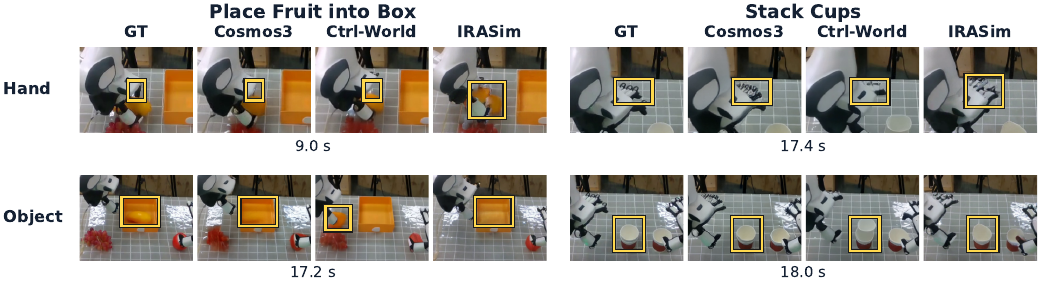}
\caption{\textbf{Real-world rollout case studies.}
Time-aligned hand (top) and object (bottom) views.}
\label{fig:real_robot_rollout_case_study}
\end{figure}
\endgroup

\section{Conclusion}
\label{sec:conclusion}

We introduced \benchmarkname{} to evaluate fine-grained control accuracy, interaction fidelity, and downstream execution utility in dexterous world models. Its full task sequences, critical interaction clips, and controlled action comparisons provide complementary views of model performance. Across six baseline models, interaction prediction improves with action conditioning, but finger-level responsiveness remains weak, and imagined video plans paired with an action decoder achieve limited task completion in simulation. Visual quality metrics alone also provide limited evidence of correct interaction outcomes. These findings identify fine-grained action grounding and the utility of video plans for downstream execution as priorities for future dexterous world models.

\clearpage
{
\small
\bibliographystyle{ieeenat_fullname}
\bibliography{main}
}

\clearpage
\maketitlesupplementary
\appendix
\section{Baseline Training and Generation Configuration}
\label{sec:app_baseline_config}
We standardize the training data, training budget, learning-rate schedule, and video specifications across baselines, as summarized in Table~\ref{tab:baseline_config}.

\begin{table}[ht]
\centering
\caption{Unified training and video configuration for the baselines.}
\label{tab:baseline_config}
\begingroup
\small
\setlength{\tabcolsep}{4pt}
\renewcommand{\arraystretch}{1.15}
\begin{tabular}{@{}>{\raggedright\arraybackslash}p{0.26\linewidth}>{\raggedright\arraybackslash}p{0.70\linewidth}@{}}
\toprule
\textbf{Setting} & \textbf{Configuration} \\
\midrule
Training budget & 100 reference epochs, each covering all valid training windows. \\
Learning rate and schedule & Base learning rate of $10^{-5}$; warmup over the first 5\% of optimizer updates, followed by cosine decay to 1\% of the initial rate. \\
Frame rate and window starts & 5 FPS, with training window starts spaced 0.2\,s apart. \\
Resolution and preprocessing & $256\times256$ conditioning images, training targets, and generated outputs; full RGB frames resized using bicubic interpolation with antialiasing. \\
Output view & A single primary camera selected by task and data variant; generation uses the corresponding view in the sample manifest. \\
Training data & Both \texttt{rand\_obj} and \texttt{rand\_full} variants of all 11 tasks: 1,980 training records. \\
Training and validation split & Episodes 0--89 for training and 90--99 for validation within each task and variant, totaling 1,980/220 records; normalization statistics computed from the training set. \\
\bottomrule
\end{tabular}
\endgroup
\end{table}

Generation uses the same frame rate, resolution, and primary camera view. Random seeds are determined by the evaluation set, sample, and generation chunk identifiers.

\section{Action Decoder Configuration and Reliability Analysis}
\label{sec:app_idm}

We implement the action decoder as an inverse dynamics model (IDM) that converts visual trajectories into robot control sequences. Training combines multi-task learning with task-specific adaptation. We first train on a mixture of original demonstrations, randomized data, perturbed trajectories, successful and failed policy rollouts, and self-play data. This mixture broadens the state--action distribution beyond demonstration trajectories and is intended to improve robustness to motion perturbations and imperfect execution states. We then fine-tune the decoder on each task's original demonstrations to improve task-specific action recovery. The multi-task stage comprises 120k training steps, followed by up to 50k task-specific steps. Within each task, the selected decoder checkpoint is held fixed across all evaluated video models.

Our formulation follows the use of current and future images to recover intervening action sequences in GR00T N1~\citep{gr00tn1} and DreamGen~\citep{dreamgen}, with an additional intermediate frame providing temporal constraints. Specifically, the decoder conditions on three frames at $t$, $t+6$, and $t+12$ to predict 12 consecutive actions. On the dataset's 30\,Hz action-recording timeline, adjacent conditioning frames are separated by six action steps, corresponding to 0.2 seconds on the video timeline and matching the 5\,fps sampling used to train the baselines. Each decoding window therefore spans two consecutive video-frame intervals while retaining the original temporal resolution of the action sequence. The intermediate frame provides an additional observation of the motion within the window, helping constrain ambiguity that remains when conditioning only on its endpoints.

We calibrate the decoder using reference videos and execute the recovered actions open-loop in simulation. On a calibration set comprising five tasks with 20 trajectories per task, the task-specific decoders achieve an average replay success rate of \textbf{93\%}, averaged over takeover at the trajectory midpoint and 60 action steps before the reference success event. This result characterizes action recovery from reference videos under intermediate-state initialization and provides a calibration reference for interpreting downstream execution metrics.

Residual execution errors may arise from three sources. First, occlusion of the hand and contact regions makes some joint states difficult to infer, while similar visual changes may correspond to different action sequences. Second, small action errors can accumulate during open-loop execution without feedback correction, affecting precise contact establishment and stable contact maintenance. Third, simulation replay itself can introduce execution deviations: in an earlier replay check, directly executing recorded ground-truth actions achieved a success rate of \textbf{93.6\%}, corresponding to a \textbf{6.4\% failure rate even without action decoding}. Execution success thus reflects the joint performance of the video model and the fixed action decoder under the specified conditions, including errors from action recovery and simulation replay.

\section{Object Trajectory Visualization}
\label{sec:app_ots_visualization}
\begin{figure}[!ht]
\centering
\begingroup
\setlength{\parskip}{0pt}
\setlength{\tabcolsep}{3pt}
\renewcommand{\arraystretch}{1.0}
\begin{tabular}{@{}>{\small\raggedright\arraybackslash}m{0.14\textwidth}*{4}{>{\centering\arraybackslash}m{0.198\textwidth}}@{}}
\toprule
\textbf{Model}\par Case OTS $\uparrow$ & \textbf{1.2\,s} & \textbf{3.6\,s} & \textbf{6.6\,s} & \textbf{7.2\,s} \\
\midrule
\textbf{GT}\par\smallskip --- & \includegraphics[width=\linewidth]{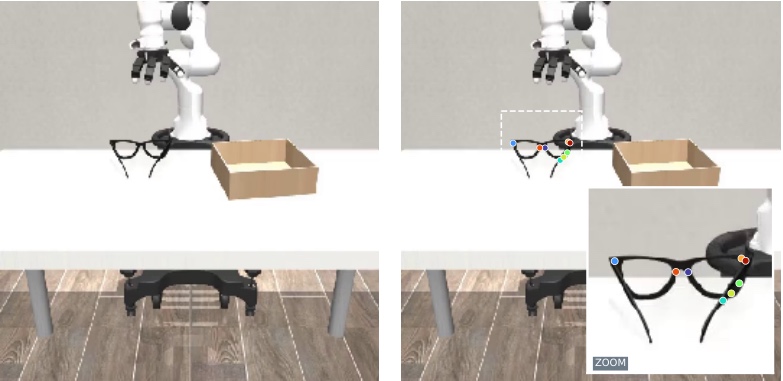} & \includegraphics[width=\linewidth]{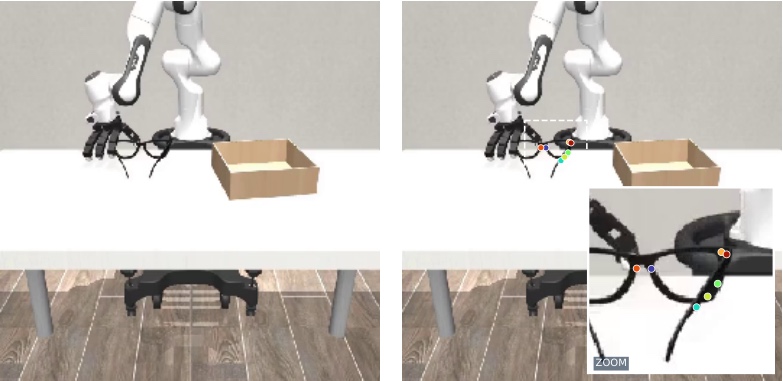} & \includegraphics[width=\linewidth]{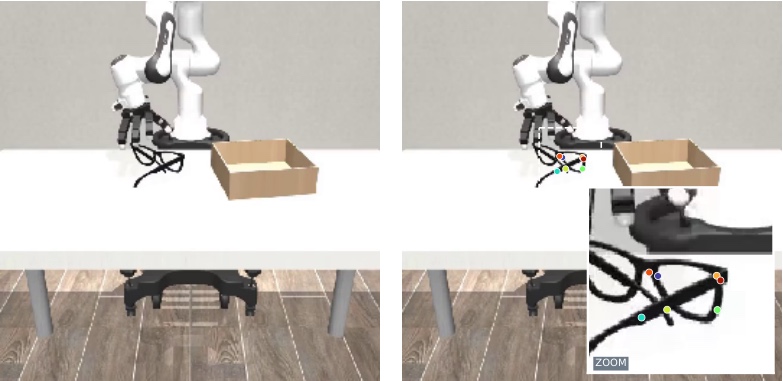} & \includegraphics[width=\linewidth]{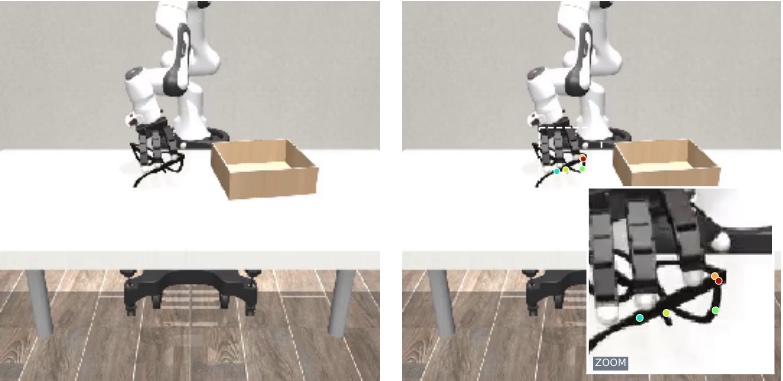} \\[4pt]
\textbf{FastWAM}\par\smallskip 0.5008 & \includegraphics[width=\linewidth]{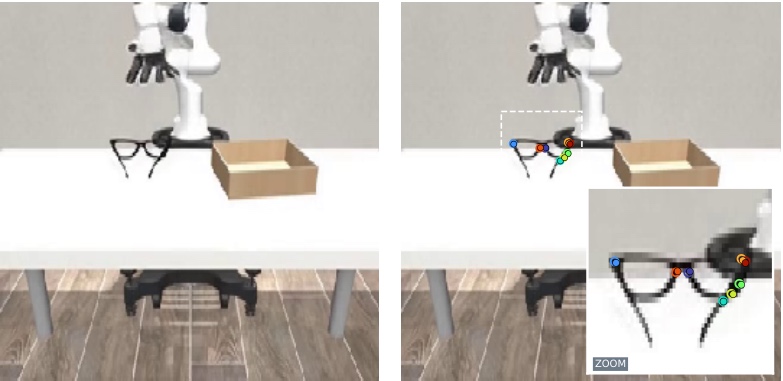} & \includegraphics[width=\linewidth]{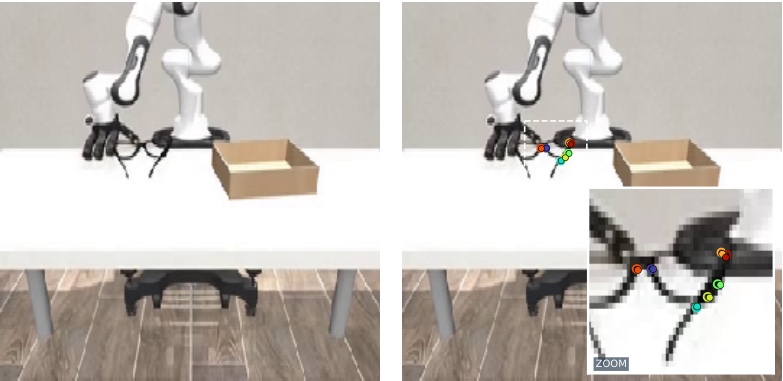} & \includegraphics[width=\linewidth]{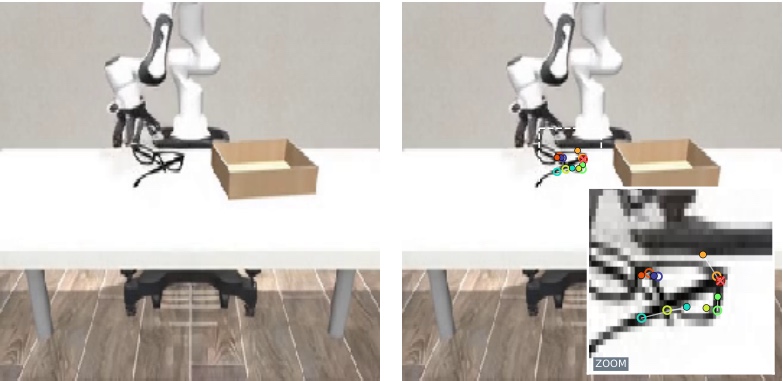} & \includegraphics[width=\linewidth]{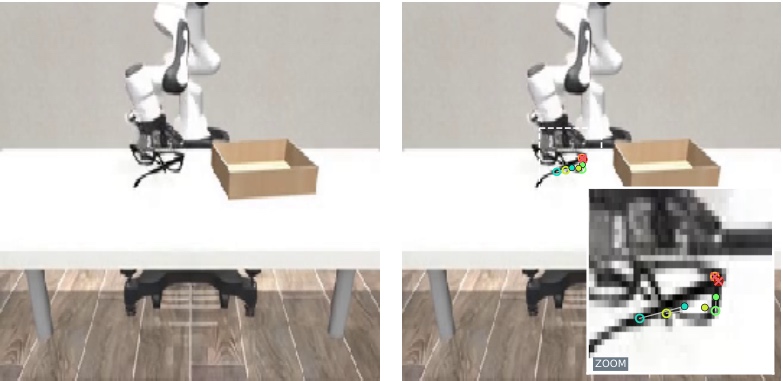} \\[4pt]
\textbf{Ctrl-World}\par\smallskip 0.5085 & \includegraphics[width=\linewidth]{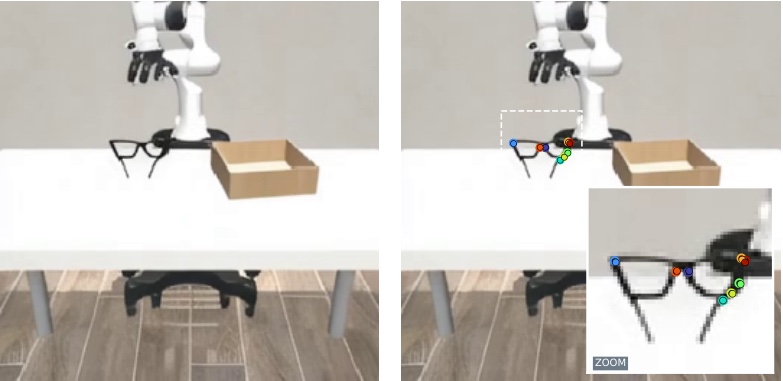} & \includegraphics[width=\linewidth]{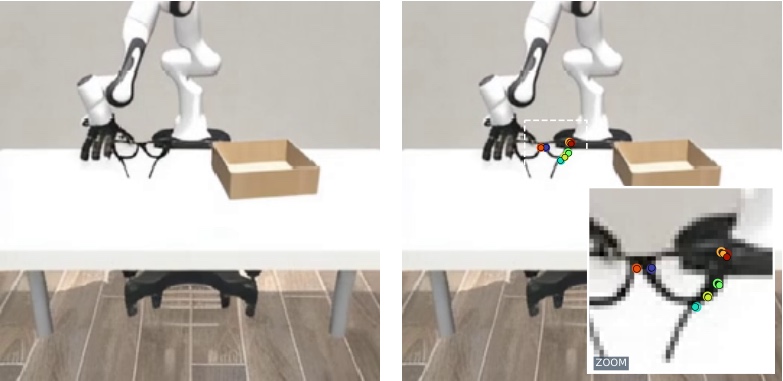} & \includegraphics[width=\linewidth]{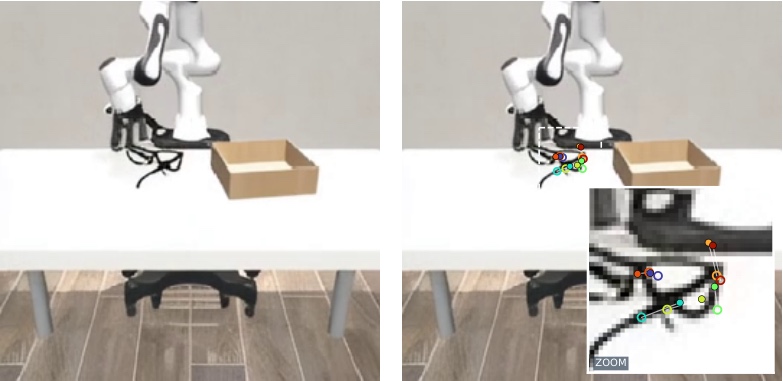} & \includegraphics[width=\linewidth]{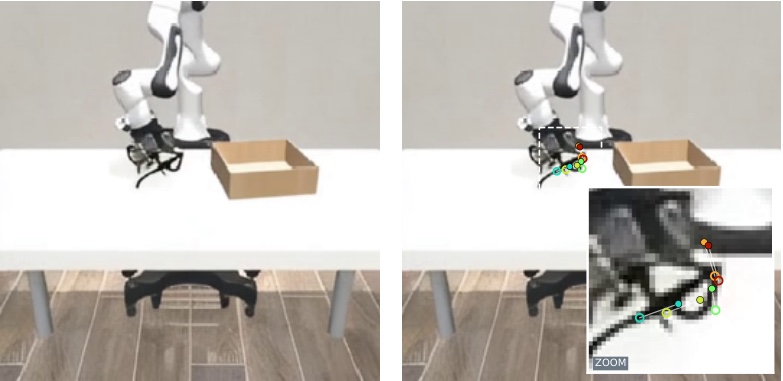} \\[4pt]
\textbf{Cosmos3}\par\smallskip 0.4351 & \includegraphics[width=\linewidth]{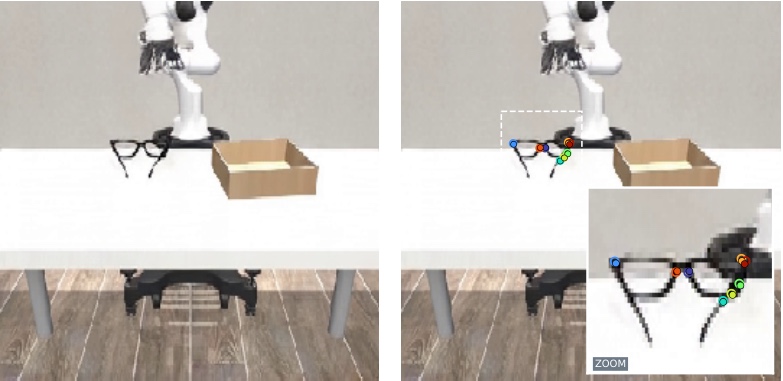} & \includegraphics[width=\linewidth]{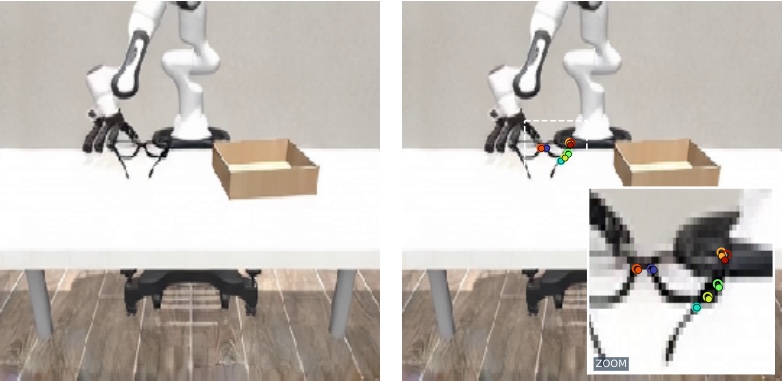} & \includegraphics[width=\linewidth]{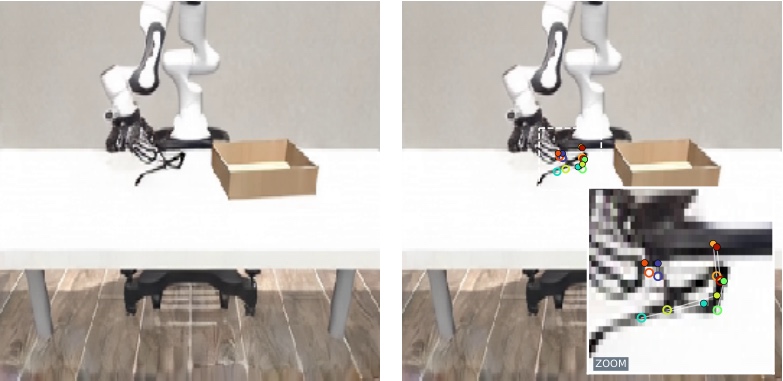} & \includegraphics[width=\linewidth]{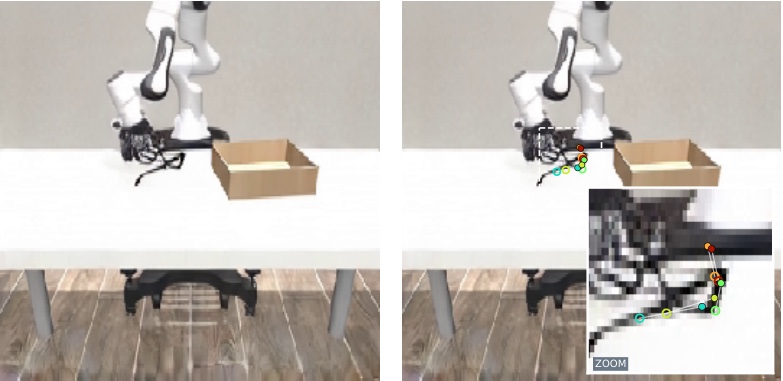} \\[4pt]
\textbf{Mask2Real-\newline WM}\par\smallskip 0.4797 & \includegraphics[width=\linewidth]{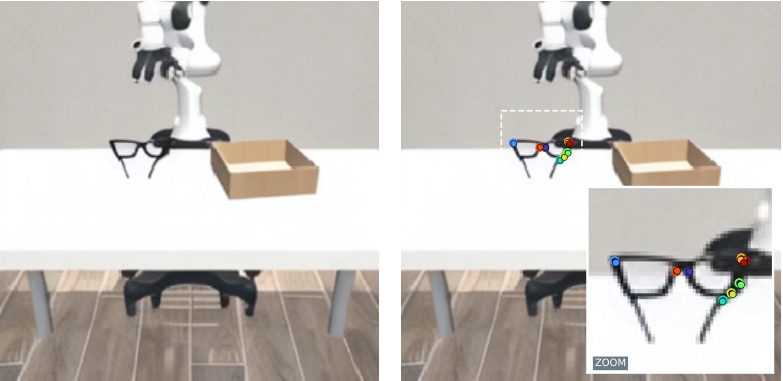} & \includegraphics[width=\linewidth]{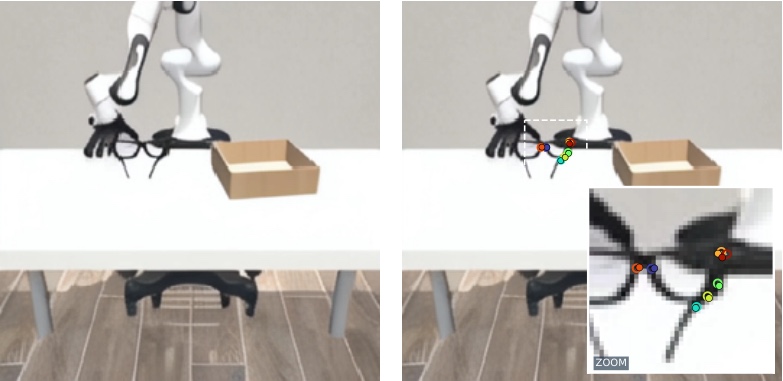} & \includegraphics[width=\linewidth]{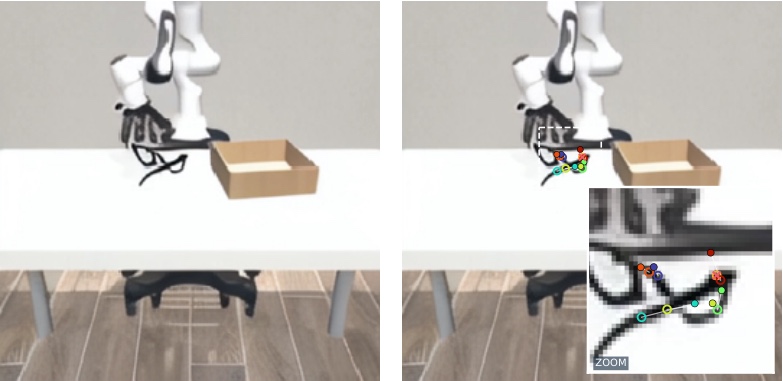} & \includegraphics[width=\linewidth]{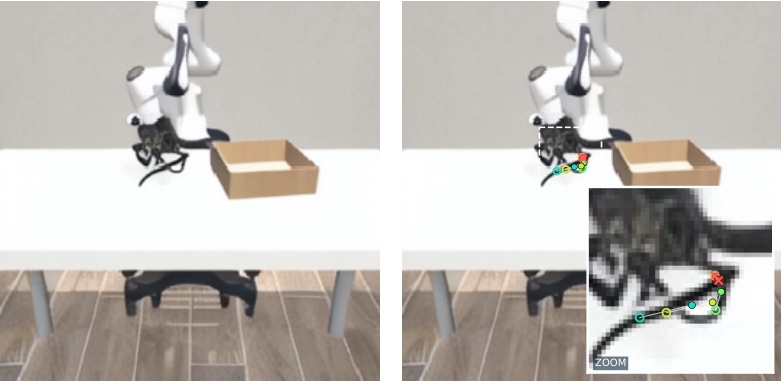} \\[4pt]
\textbf{IRASim}\par\smallskip 0.4464 & \includegraphics[width=\linewidth]{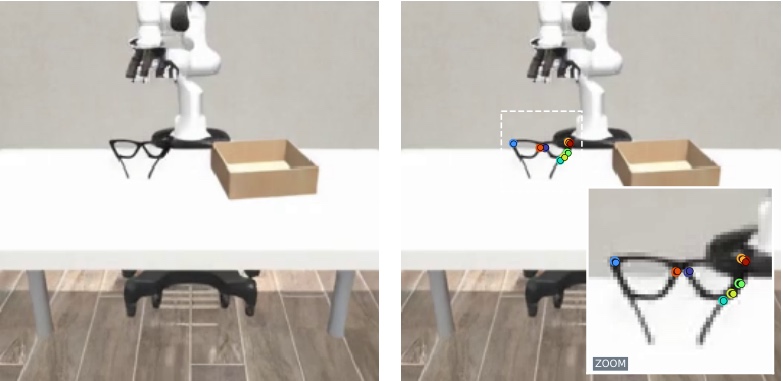} & \includegraphics[width=\linewidth]{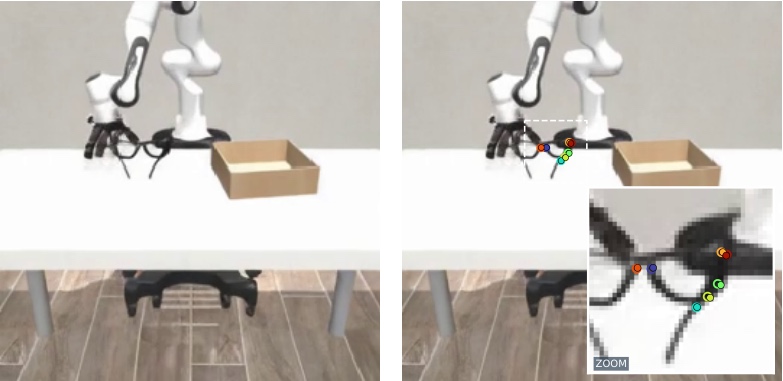} & \includegraphics[width=\linewidth]{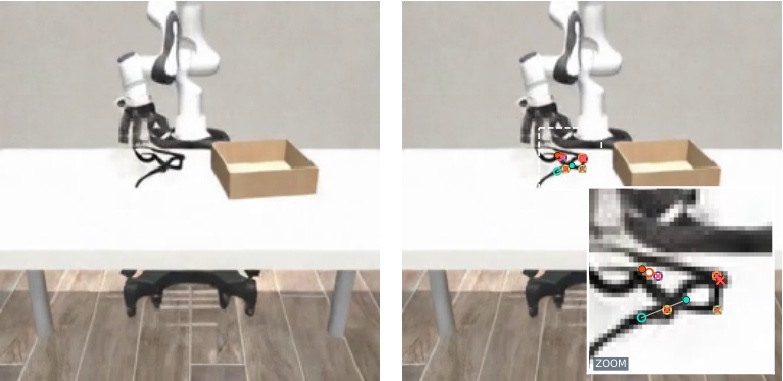} & \includegraphics[width=\linewidth]{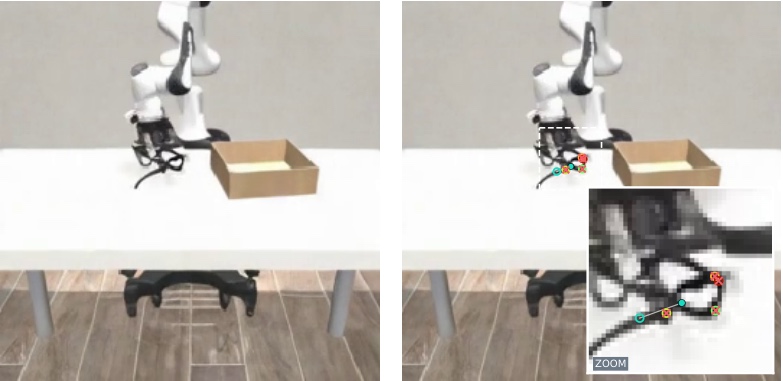} \\[4pt]
\textbf{Wan}\par\smallskip 0.4063 & \includegraphics[width=\linewidth]{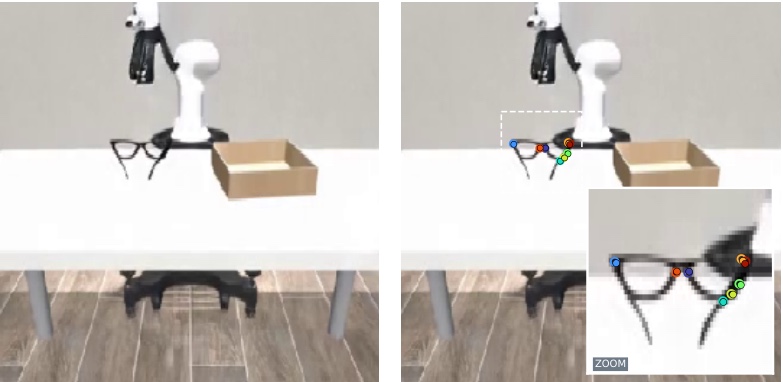} & \includegraphics[width=\linewidth]{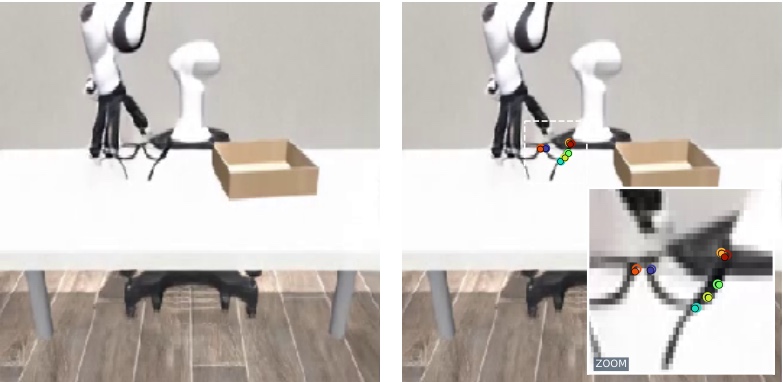} & \includegraphics[width=\linewidth]{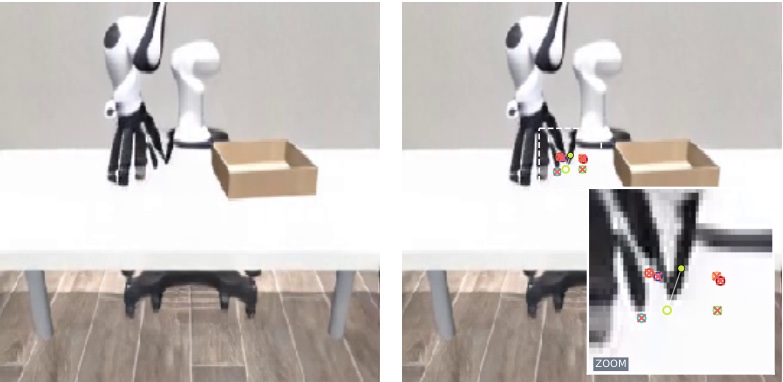} & \includegraphics[width=\linewidth]{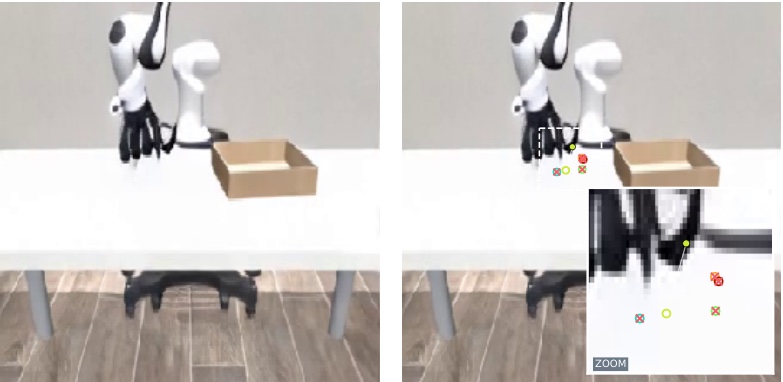} \\[4pt]
\bottomrule
\end{tabular}
\endgroup
\caption{\textbf{Object trajectory comparison on fold glasses.} Each time cell pairs the unannotated original frame (left) with the tracked-point overlay and a magnified object inset (right). Colors identify the same reference points across rows and times. In model rows, filled dots denote tracked positions, hollow circles denote the GT reference positions, white lines connect corresponding positions, and red crosses mark missing tracks. The GT row shows reference points. OTS values describe the full case, rather than only these four frames; GT is shown as the reference.}
\label{fig:ots_fold_glasses}
\end{figure}

Figure~\ref{fig:ots_fold_glasses} compares the ground-truth video and six baselines on the same \emph{fold glasses} case at four matched timestamps, together with each model's case-level Object Trajectory Score (OTS).

\clearpage
\section{Generalization Results}
\label{sec:app_generalization}
Figures~\ref{fig:gen_appearance}--\ref{fig:gen_task} show GT and six baselines for appearance, object, and task generalization. Each row presents matched timestamps, the original scores, and a brief summary of the judge's findings. Aggregate results are reported in Table~\ref{tab:generalization_results}.

\newlength{\genrowheight}
\newcommand{\gencell}[1]{\parbox[c][\genrowheight][c]{\linewidth}{\setlength{\parskip}{0pt}\raggedright #1}}
\newcommand{\genframe}[1]{\gencell{\includegraphics[width=\linewidth]{#1}}}

\begin{figure}[!ht]
\centering
\begingroup
\setlength{\genrowheight}{70pt}
\fontsize{10.5}{12.5}\selectfont
\setlength{\tabcolsep}{1pt}
\renewcommand{\arraystretch}{1.0}
\begin{tabular}{@{}>{\fontsize{9}{11}\selectfont\raggedright\arraybackslash}m{0.10\textwidth}*{5}{m{0.122\textwidth}}>{\normalsize\raggedright\arraybackslash}m{0.25\textwidth}@{}}
\multicolumn{7}{@{}l}{\textbf{Appearance generalization: texture variation}} \\[5pt]
\toprule
\textbf{Model} & \centering 0\,s (input) & \centering 2\,s & \centering 8\,s & \centering 16\,s & \centering 23.2\,s & \textbf{Judge summary} \\
\midrule
\gencell{\textbf{GT}\par\smallskip $G=5$, $C=0$} & \genframe{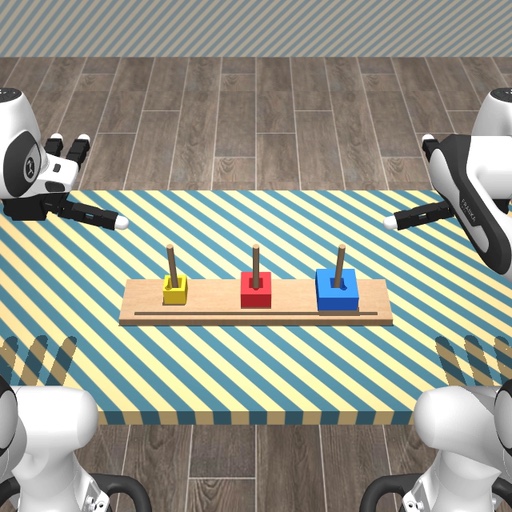} & \genframe{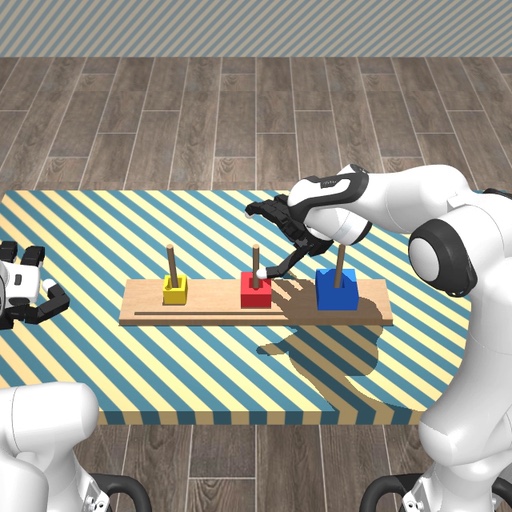} & \genframe{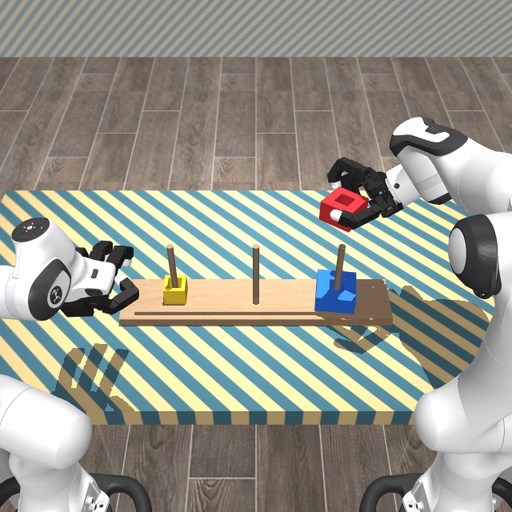} & \genframe{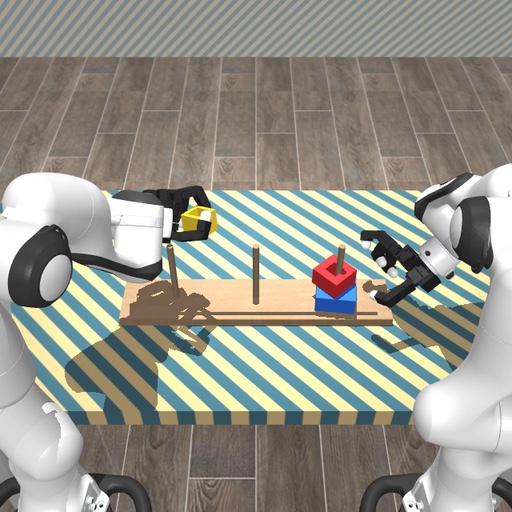} & \genframe{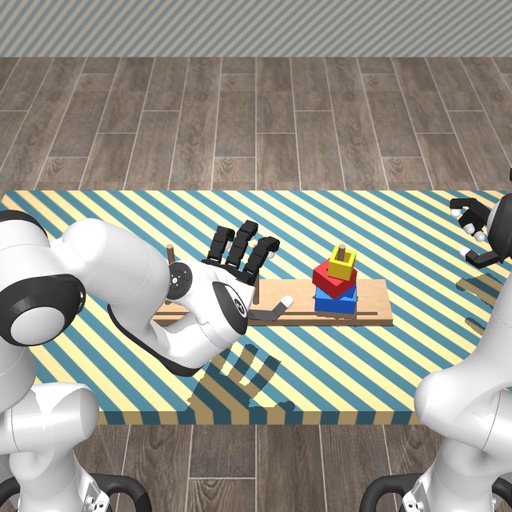} & \gencell{Tabletop and wall stripes persist throughout; robot hands and manipulated objects remain coherent.} \\
\gencell{\textbf{Wan}\par\smallskip $G=4$, $C=0$} & \genframe{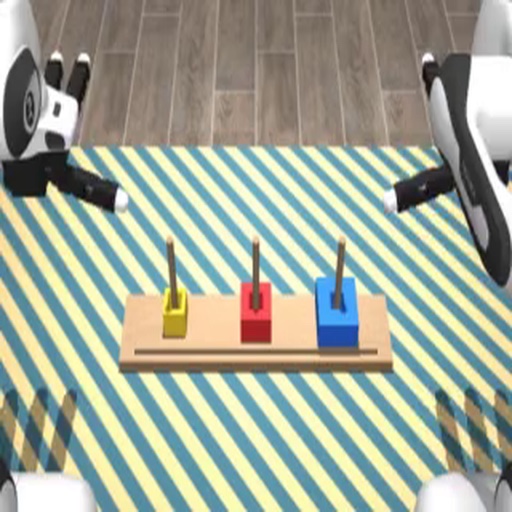} & \genframe{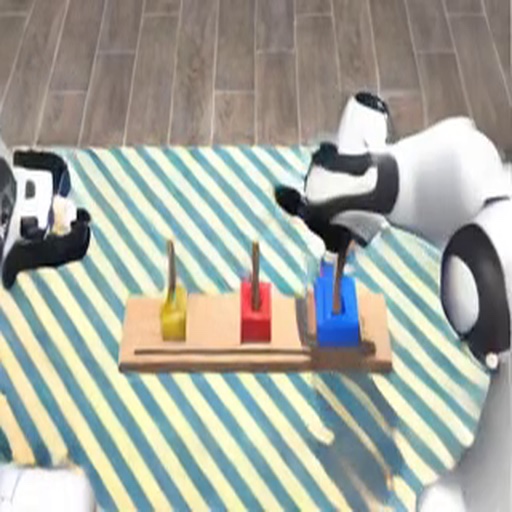} & \genframe{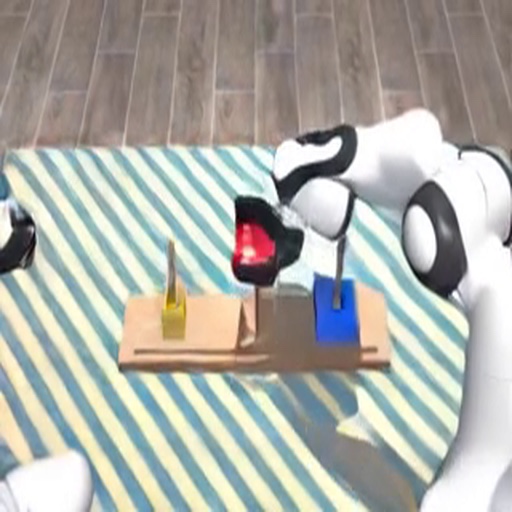} & \genframe{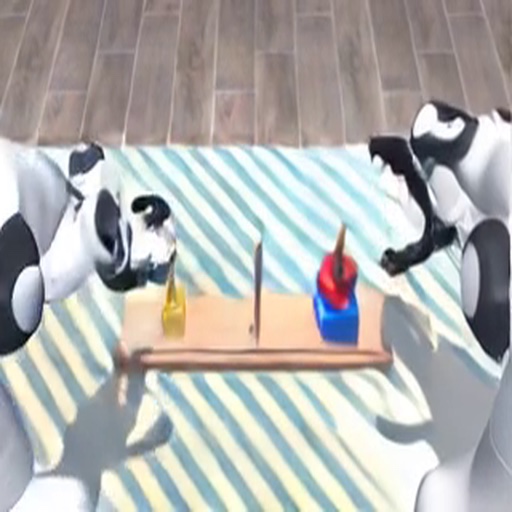} & \genframe{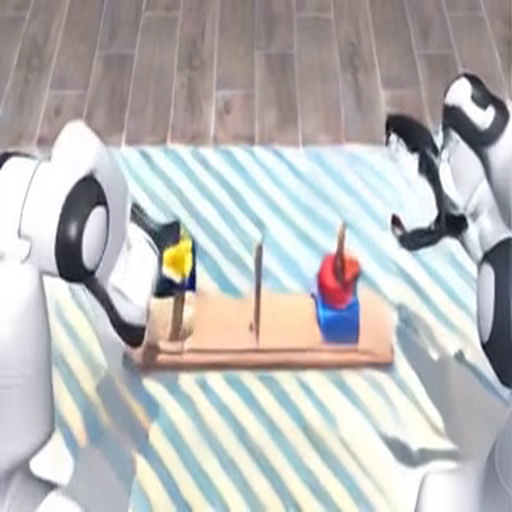} & \gencell{Stripes remain recognizable with minor waviness and softened detail. Hands retain connected structures; no severe collapse is observed.} \\
\gencell{\textbf{IRASim}\par\smallskip $G=1$, $C=0$} & \genframe{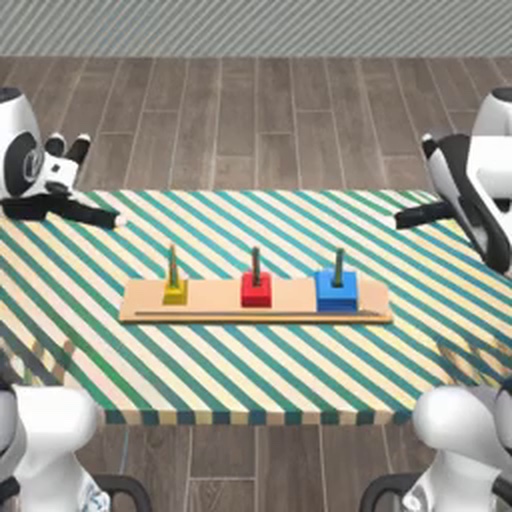} & \genframe{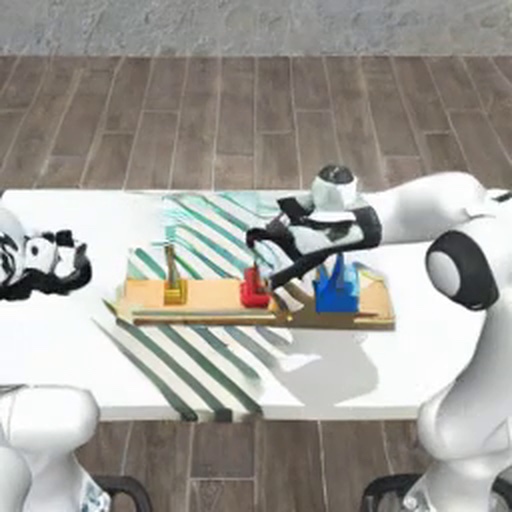} & \genframe{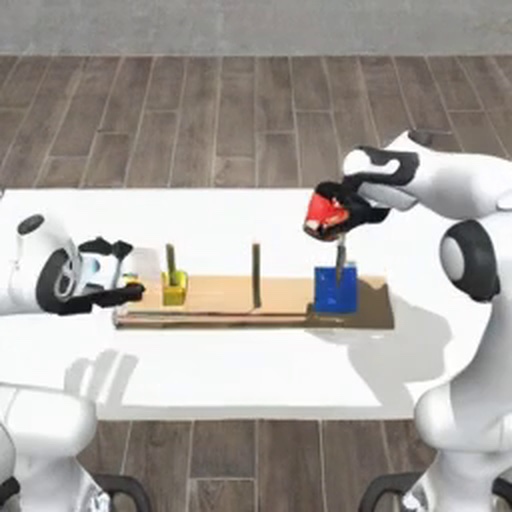} & \genframe{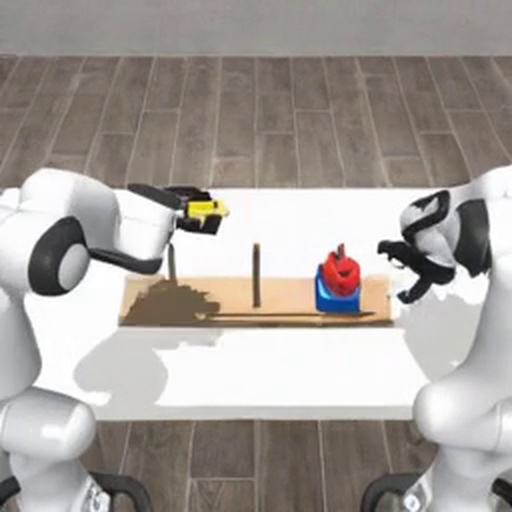} & \genframe{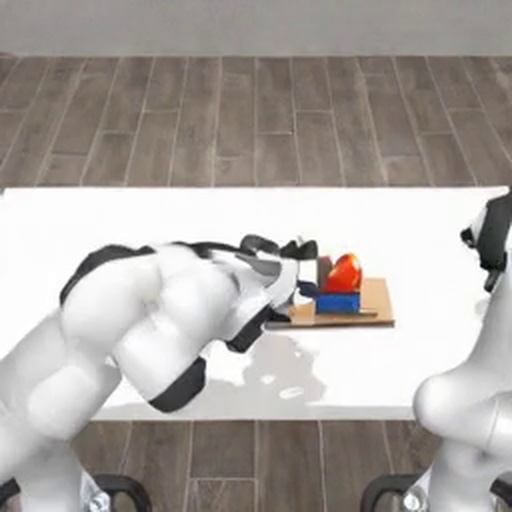} & \gencell{Tabletop and wall patterns are almost entirely erased, leaving plain surfaces. Robot articulation and scene structure remain coherent.} \\
\gencell{\textbf{Ctrl-World}\par\smallskip $G=3$, $C=0$} & \genframe{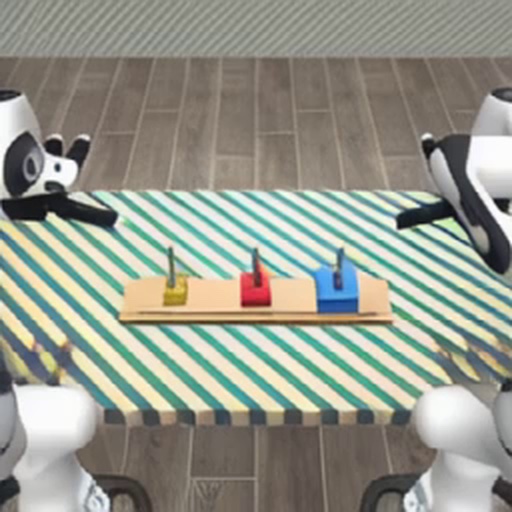} & \genframe{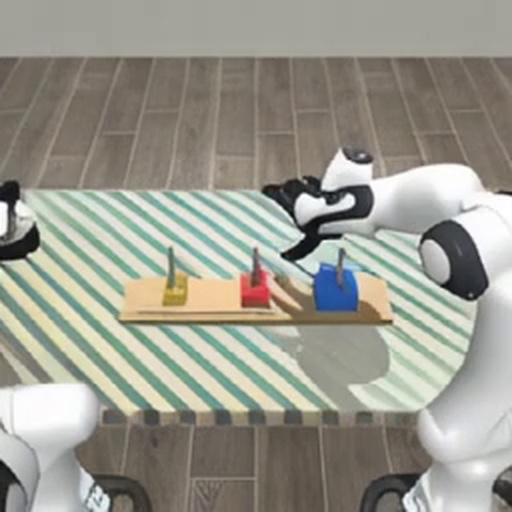} & \genframe{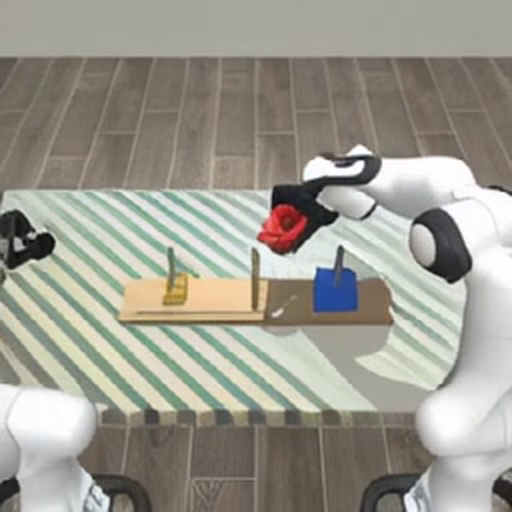} & \genframe{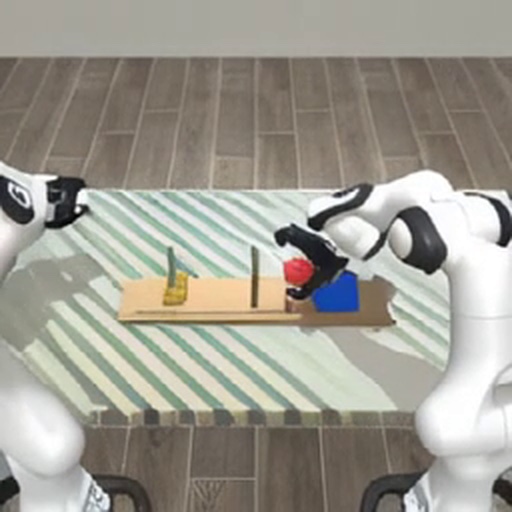} & \genframe{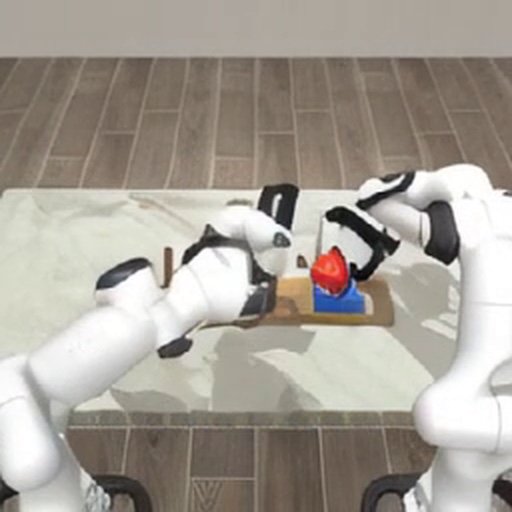} & \gencell{Stripes change color and largely fade in late frames; the wall becomes plain. Arms, hands, and scene layout remain coherent.} \\
\gencell{\textbf{FastWAM}\par\smallskip $G=1$, $C=1$} & \genframe{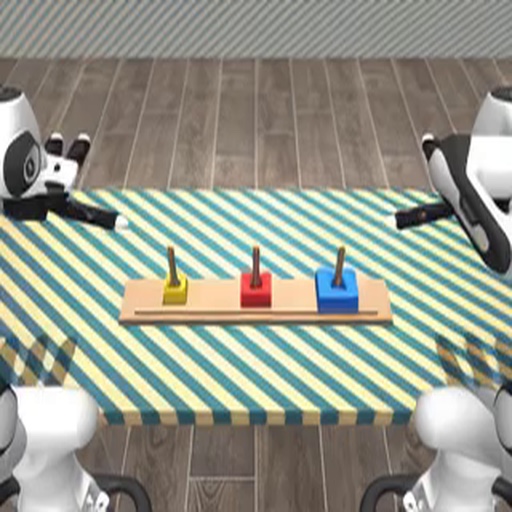} & \genframe{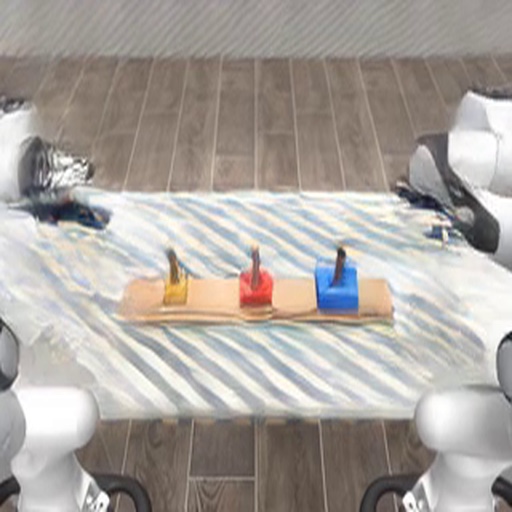} & \genframe{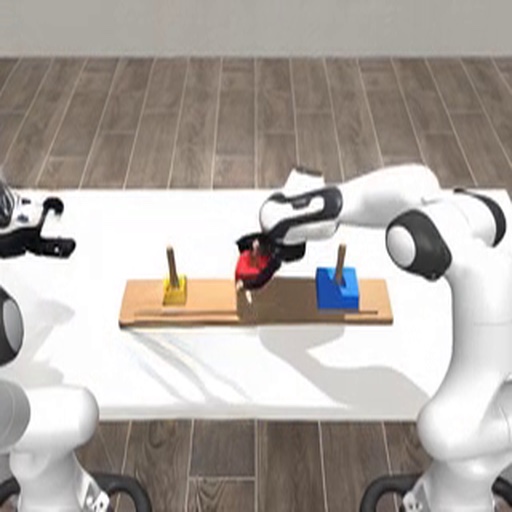} & \genframe{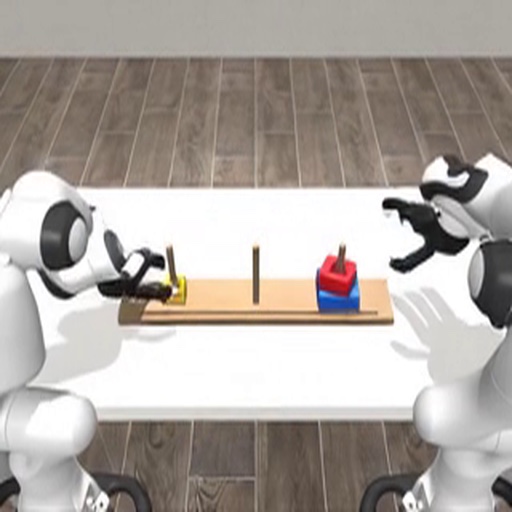} & \genframe{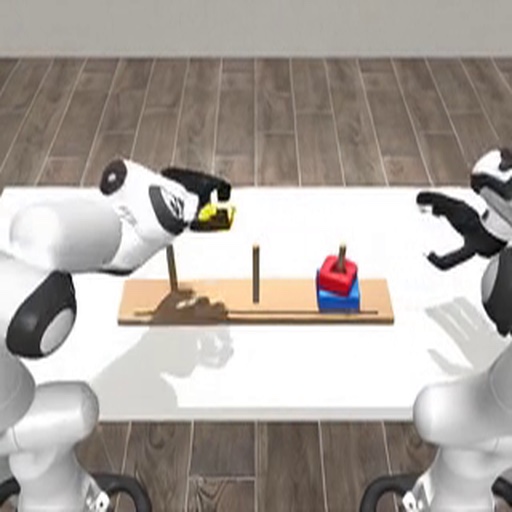} & \gencell{Reference patterns rapidly wash out into plain surfaces. The early wrist and hand break into tangled strands and overlapping fragments.} \\
\gencell{\textbf{Cosmos3}\par\smallskip $G=1$, $C=1$} & \genframe{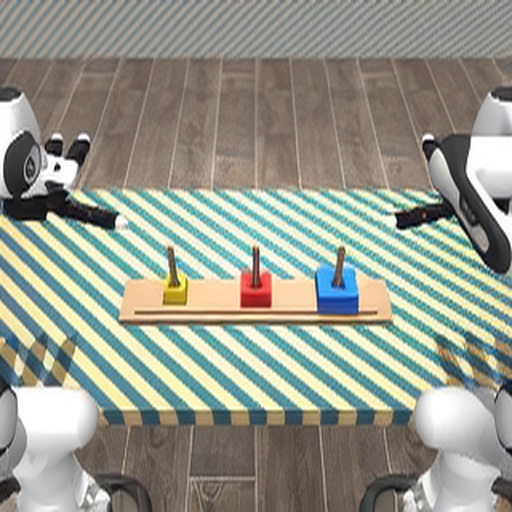} & \genframe{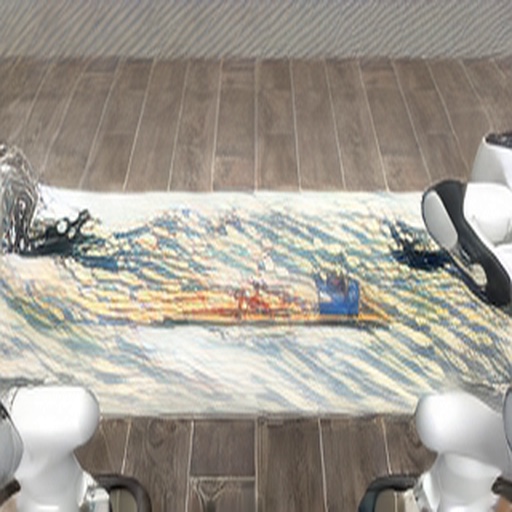} & \genframe{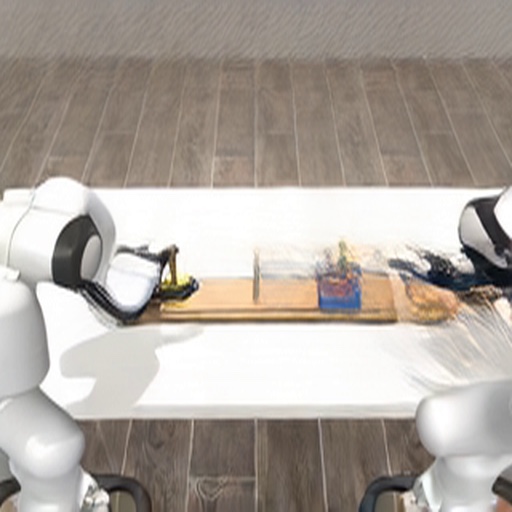} & \genframe{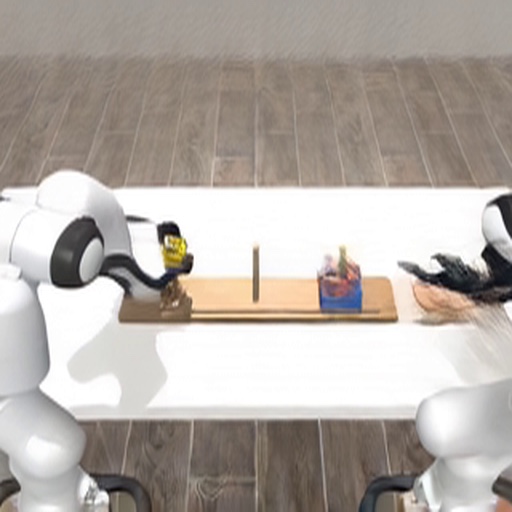} & \genframe{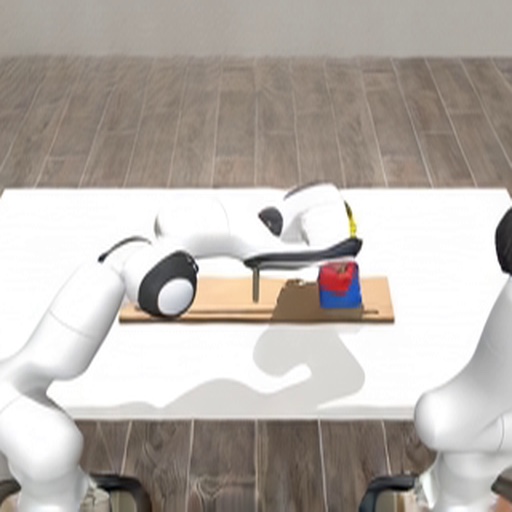} & \gencell{Tabletop and wall patterns are almost entirely erased. Early hand structures dissolve into branching strands, losing coherent palms and fingers.} \\
\gencell{\textbf{Mask2Real-\newline WM}\par\smallskip $G=3$, $C=0$} & \genframe{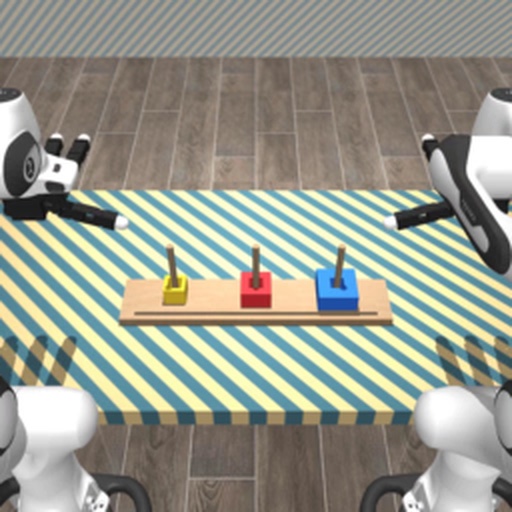} & \genframe{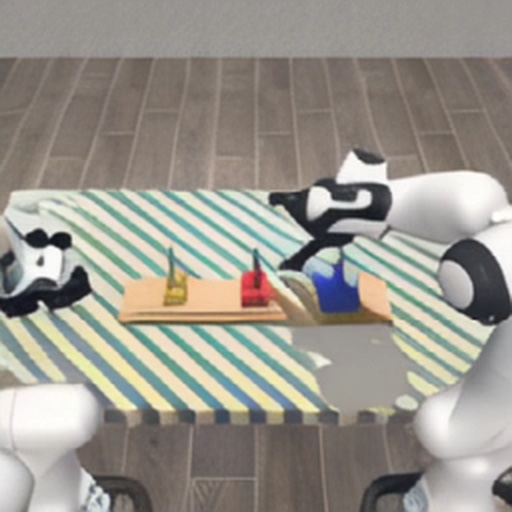} & \genframe{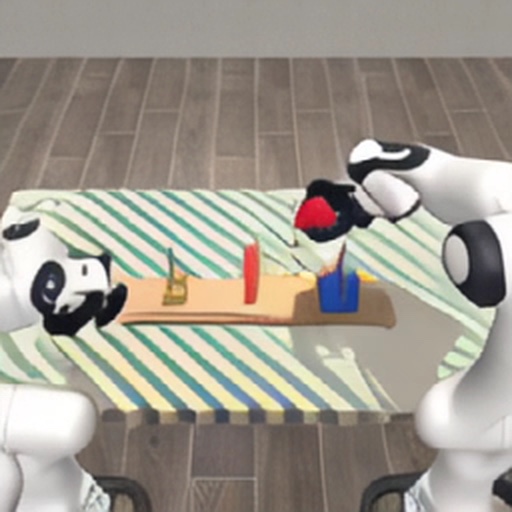} & \genframe{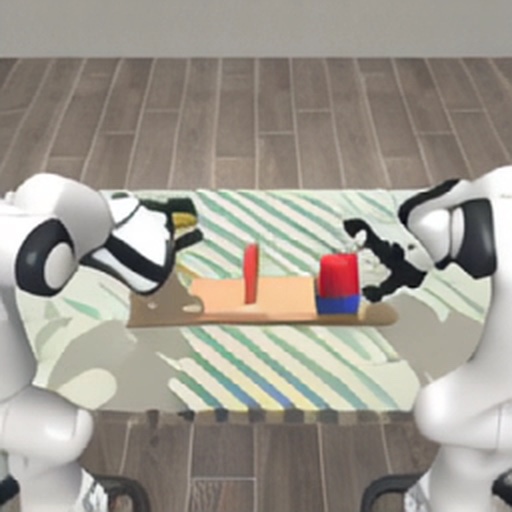} & \genframe{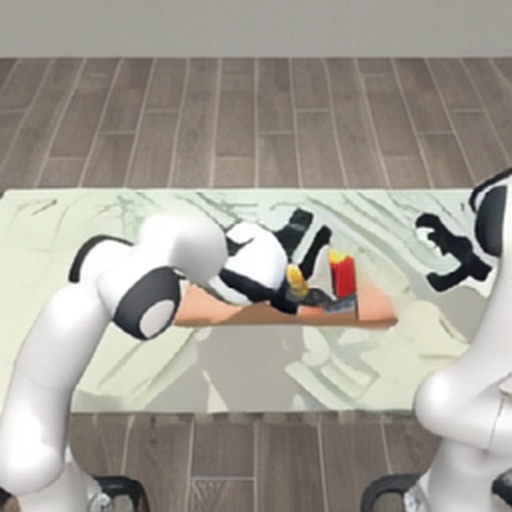} & \gencell{Wall stripes vanish; tabletop stripes fade to a pale surface. Arms and grippers keep coherent, connected shapes.} \\
\bottomrule
\end{tabular}
\endgroup
\caption{\textbf{Texture variation in bimanual Tower of Hanoi.} Texture loss can occur without structural collapse, as shown by IRASim. $G$: texture-preservation score (1--5, higher is better); $C$: visual collapse (0: absent, 1: present). Reasons summarize the original judgments; scores refer to the full judged sequence.}
\label{fig:gen_appearance}
\end{figure}
\clearpage
\begin{figure}[!ht]
\centering
\begingroup
\setlength{\genrowheight}{79pt}
\fontsize{10.5}{12.5}\selectfont
\setlength{\tabcolsep}{1pt}
\renewcommand{\arraystretch}{1.0}
\begin{tabular}{@{}>{\fontsize{9}{11}\selectfont\raggedright\arraybackslash}m{0.10\textwidth}*{5}{m{0.122\textwidth}}>{\normalsize\raggedright\arraybackslash}m{0.25\textwidth}@{}}
\multicolumn{7}{@{}l}{\textbf{Object generalization: replacement peg and socket}} \\[5pt]
\toprule
\textbf{Model} & \centering 0\,s (input) & \centering 4\,s & \centering 12\,s & \centering 24\,s & \centering 35.6\,s & \textbf{Judge summary} \\
\midrule
\gencell{\textbf{GT}\par\smallskip $G=5$, $C=0$} & \genframe{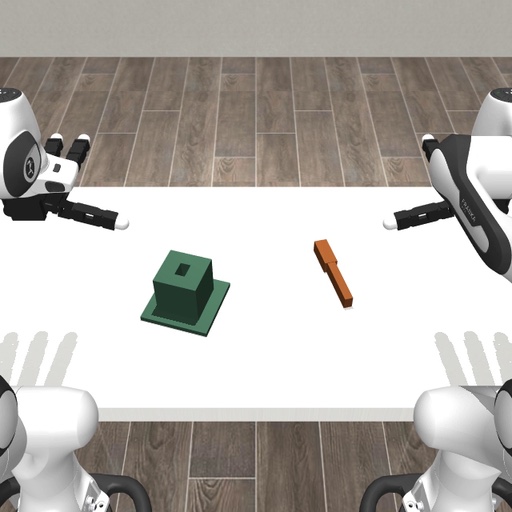} & \genframe{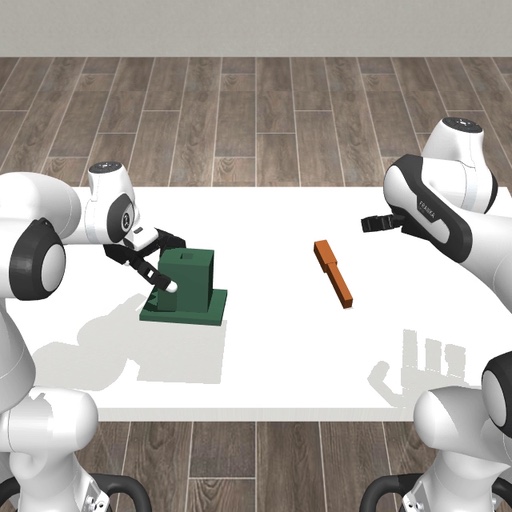} & \genframe{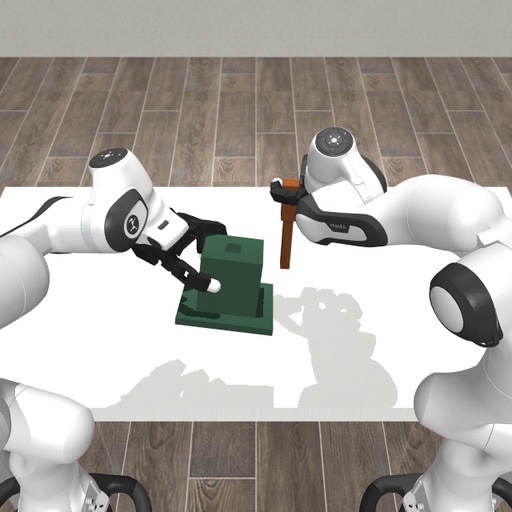} & \genframe{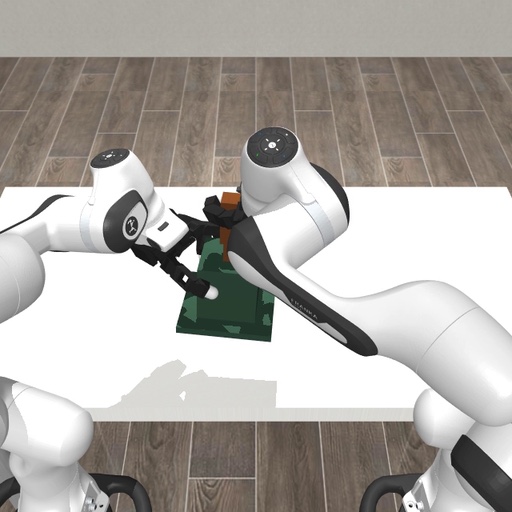} & \genframe{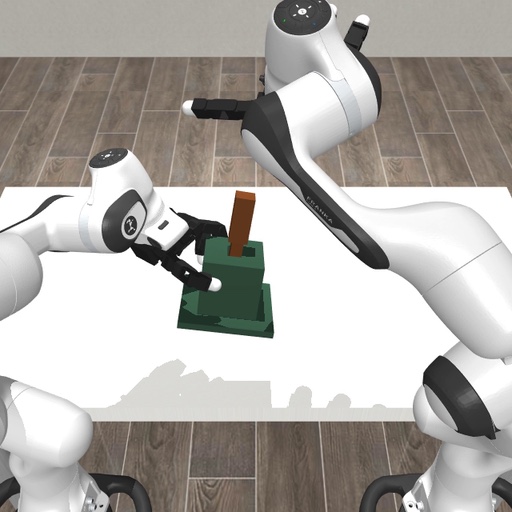} & \gencell{The brown rectangular peg and green stepped socket remain identifiable through insertion and hand occlusion, with coherent final shapes.} \\
\gencell{\textbf{Wan}\par\smallskip $G=4$, $C=1$} & \genframe{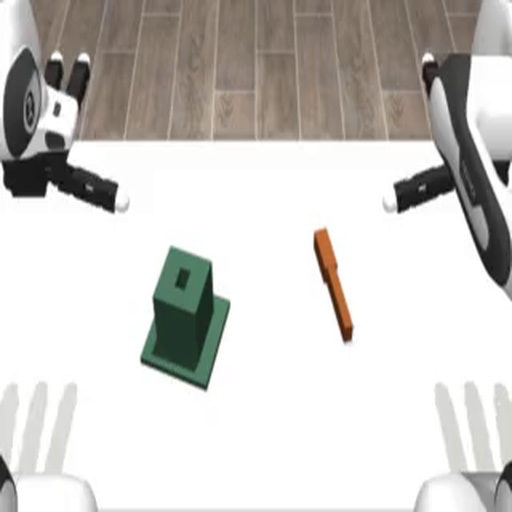} & \genframe{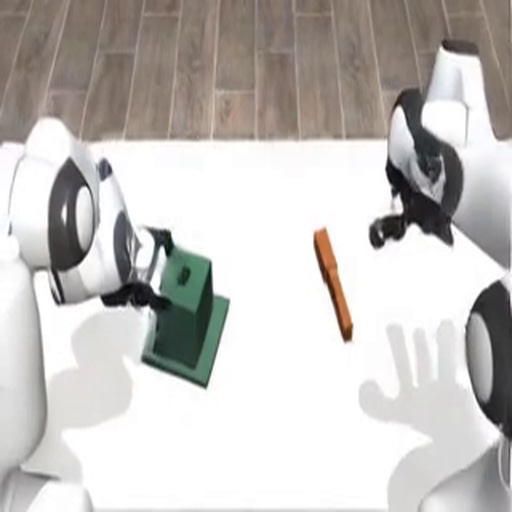} & \genframe{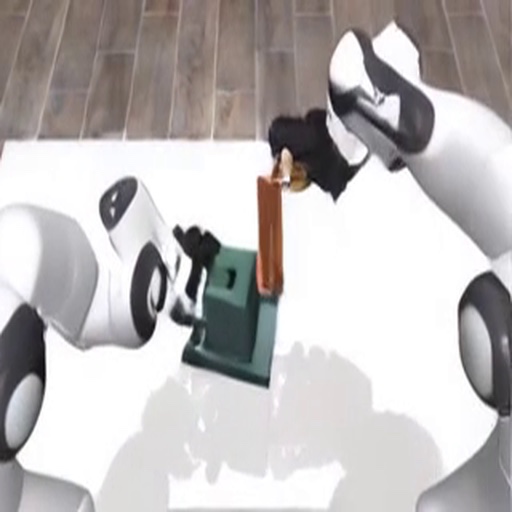} & \genframe{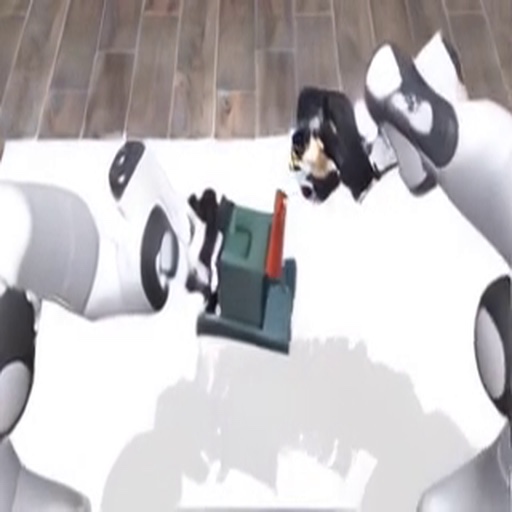} & \genframe{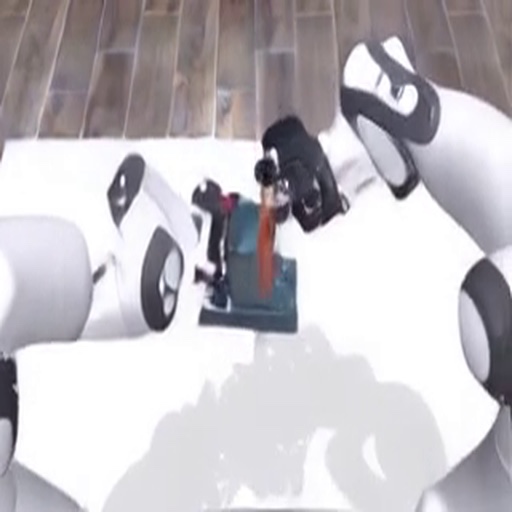} & \gencell{Both targets remain recognizable despite minor color drift. The right fingers develop irregular overlapping forms and strand-like connections near the end.} \\
\gencell{\textbf{IRASim}\par\smallskip $G=2$, $C=1$} & \genframe{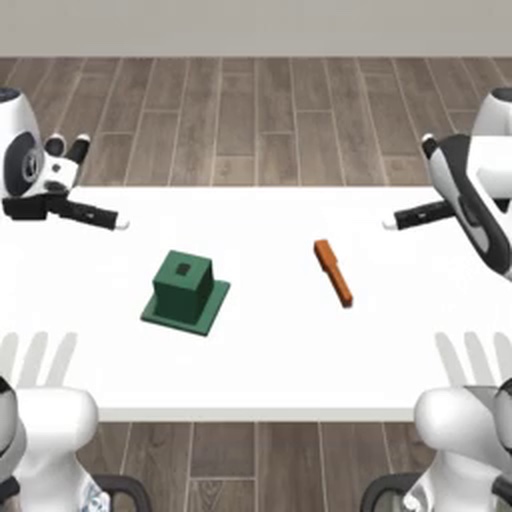} & \genframe{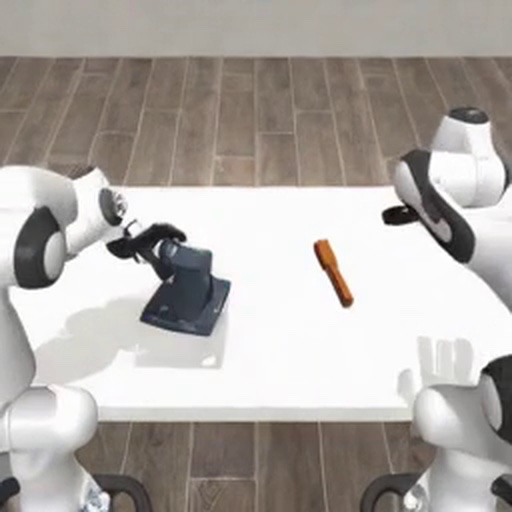} & \genframe{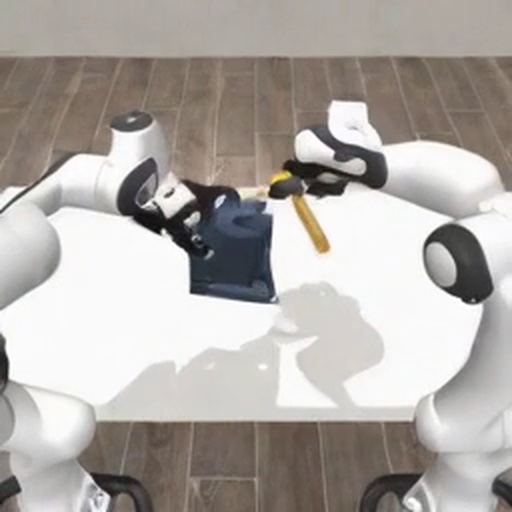} & \genframe{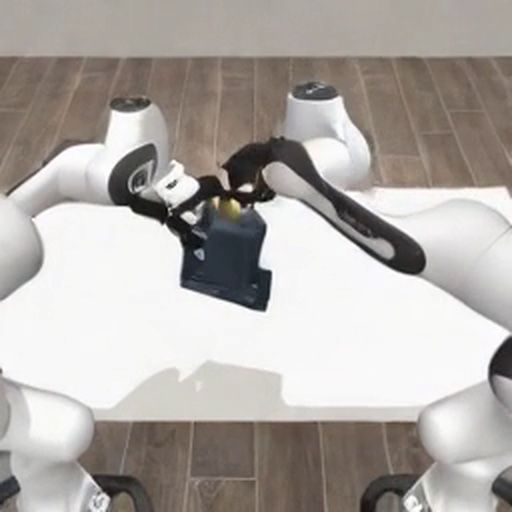} & \genframe{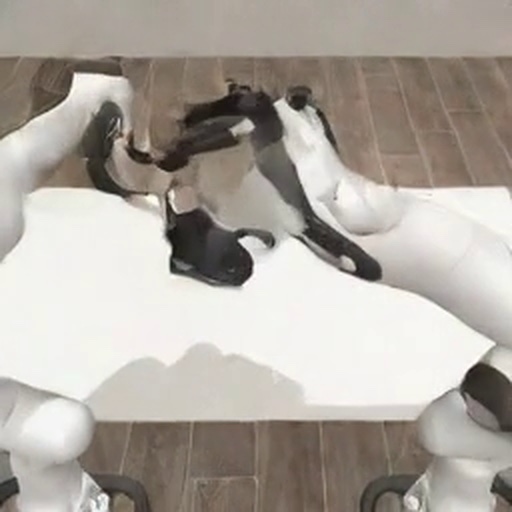} & \gencell{Target colors drift; the socket becomes a rounded hanging mass. Late peg identity is unsupported; hand--object boundaries stretch and merge.} \\
\gencell{\textbf{Ctrl-World}\par\smallskip $G=3$, $C=1$} & \genframe{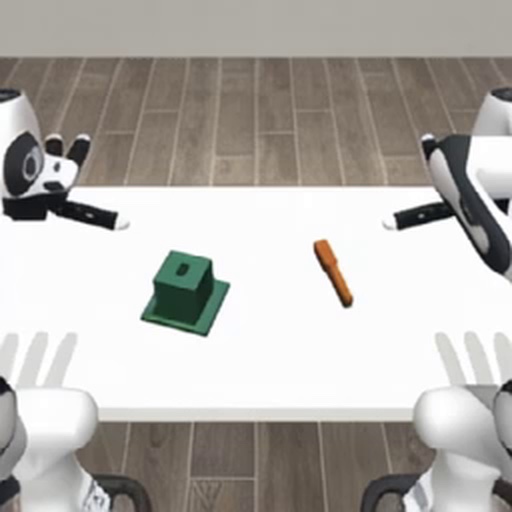} & \genframe{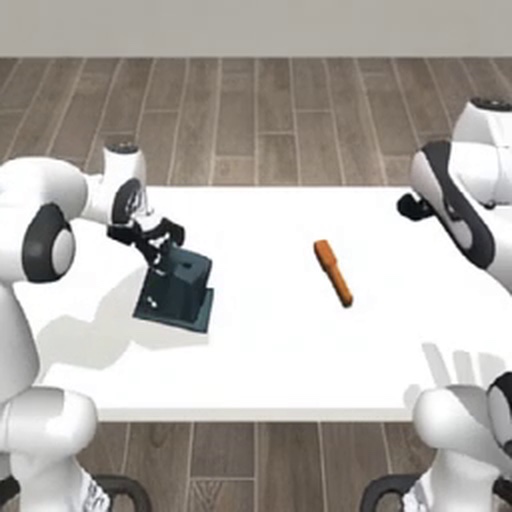} & \genframe{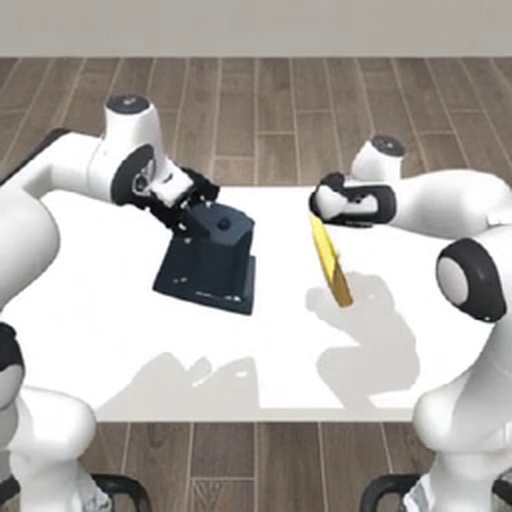} & \genframe{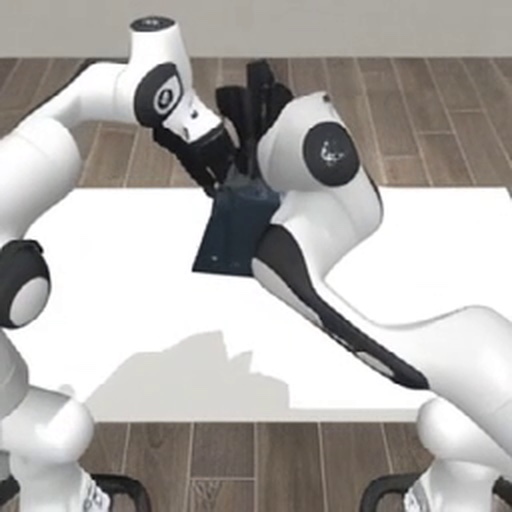} & \genframe{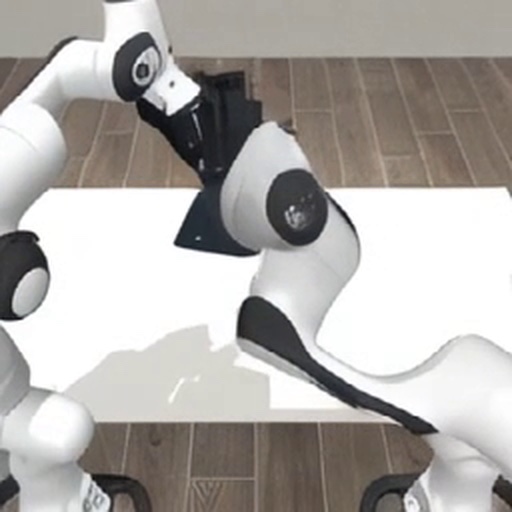} & \gencell{The socket turns blue-black and the peg yellow; late occlusion remains supported. The wrist--forearm connection deforms into a flattened, sheet-like band.} \\
\gencell{\textbf{FastWAM}\par\smallskip $G=3$, $C=0$} & \genframe{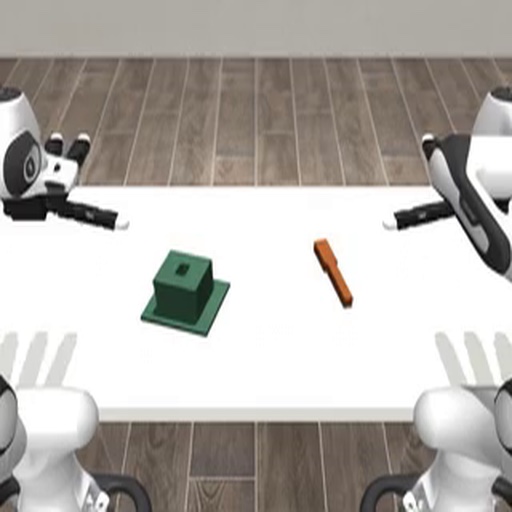} & \genframe{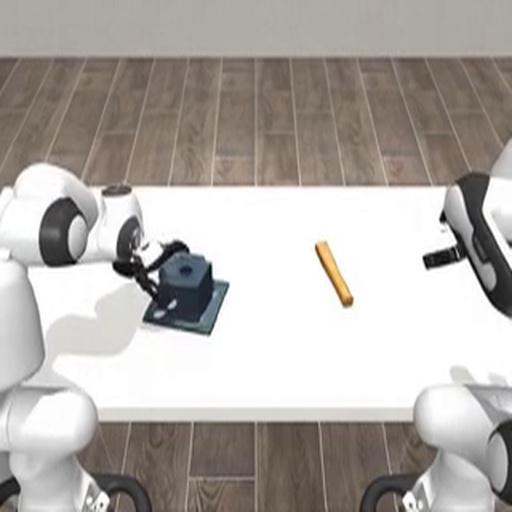} & \genframe{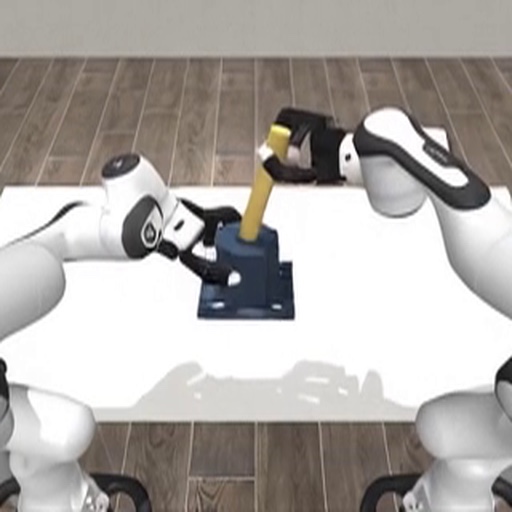} & \genframe{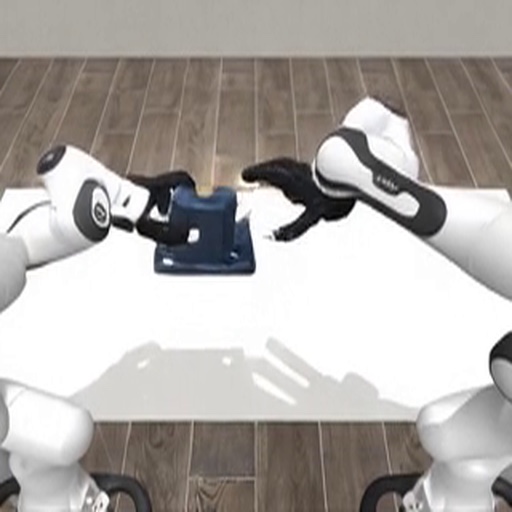} & \genframe{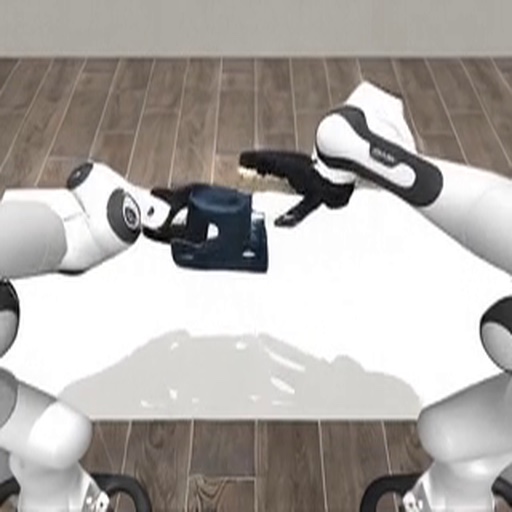} & \gencell{The socket becomes blue-gray, its opening rounder, and the peg gold. Insertion supports later occlusion; mechanical structures remain coherent.} \\
\gencell{\textbf{Cosmos3}\par\smallskip $G=2$, $C=1$} & \genframe{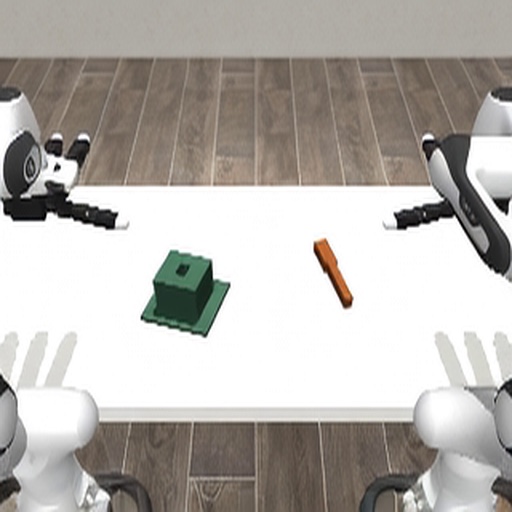} & \genframe{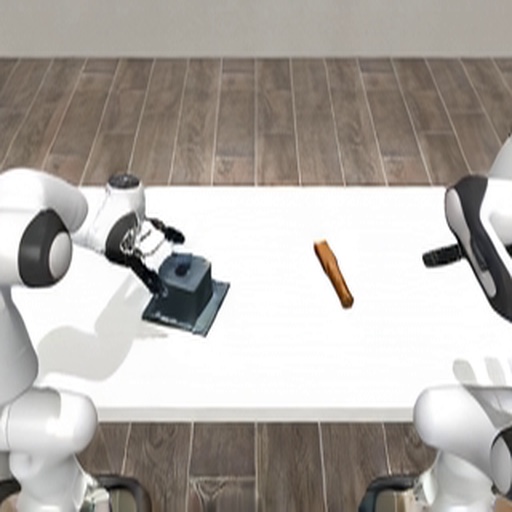} & \genframe{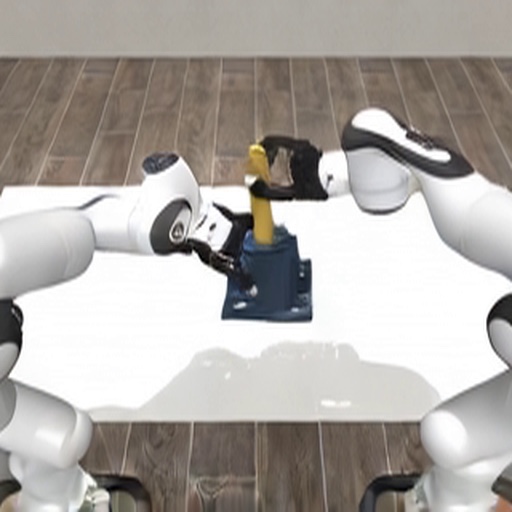} & \genframe{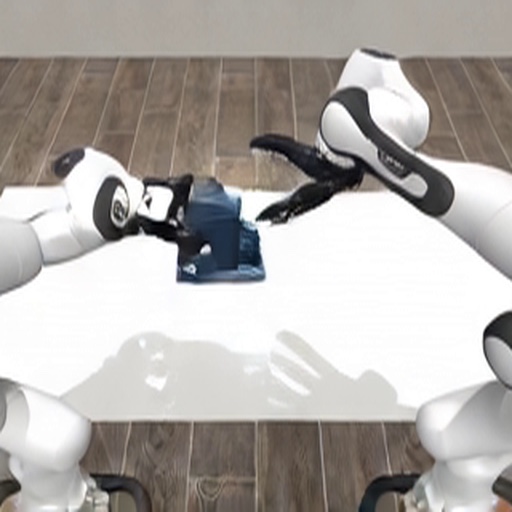} & \genframe{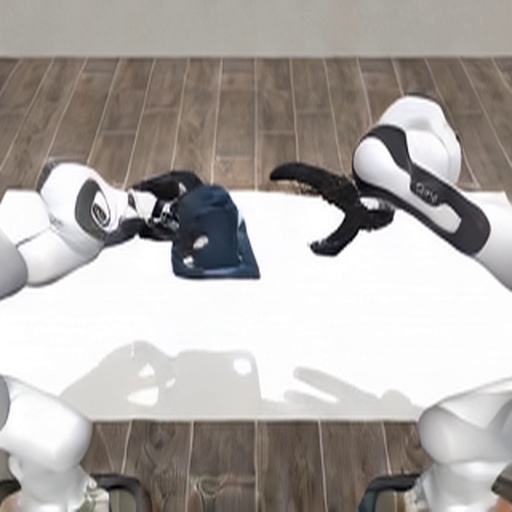} & \gencell{The peg loses its bounded shaft appearance and lacks supported late preservation. The socket collapses into an irregular shell with distorted walls and opening.} \\
\gencell{\textbf{Mask2Real-\newline WM}\par\smallskip $G=2$, $C=1$} & \genframe{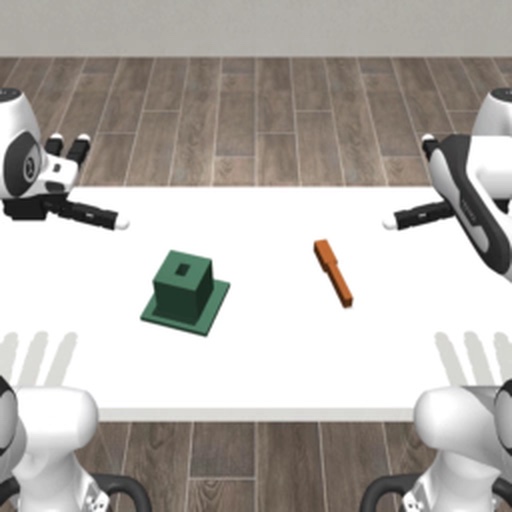} & \genframe{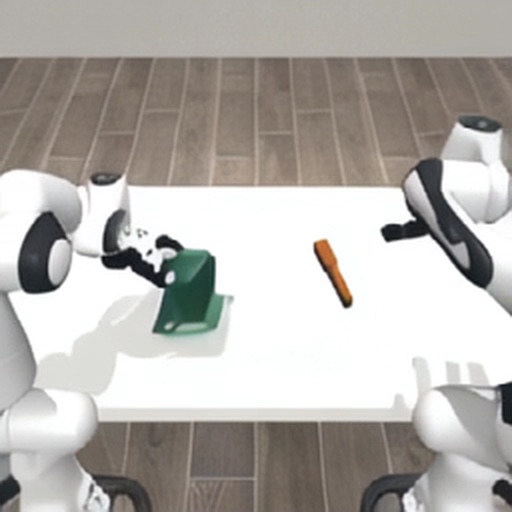} & \genframe{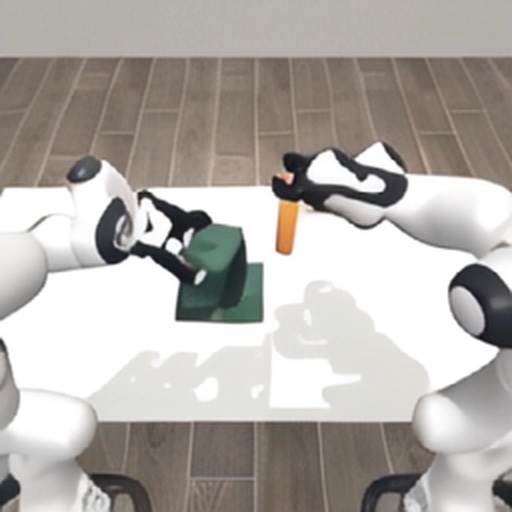} & \genframe{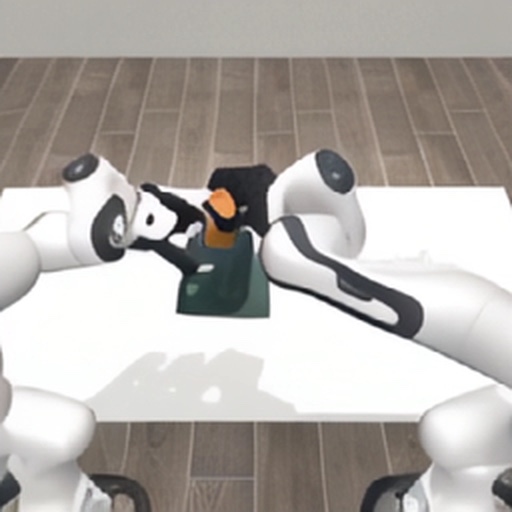} & \genframe{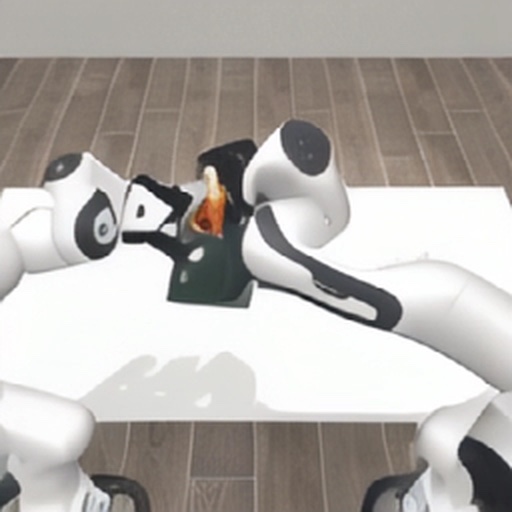} & \gencell{The socket becomes rounder. The exposed late peg develops irregular folds and pale branches, destroying its coherent shaft and clean hand--object boundary.} \\
\bottomrule
\end{tabular}
\endgroup
\caption{\textbf{Brown rectangular peg and dark green socket.} Target preservation and robot integrity can differ: Wan retains recognizable targets despite hand breakdown. $G$: object-preservation score; $C$: visual collapse. Reasons summarize the original judgments; scores refer to the full judged sequence.}
\label{fig:gen_object}
\end{figure}
\clearpage
\begin{figure}[!ht]
\centering
\begingroup
\setlength{\genrowheight}{79pt}
\fontsize{10.5}{12.5}\selectfont
\setlength{\tabcolsep}{1pt}
\renewcommand{\arraystretch}{1.0}
\begin{tabular}{@{}>{\fontsize{9}{11}\selectfont\raggedright\arraybackslash}m{0.10\textwidth}*{5}{m{0.122\textwidth}}>{\normalsize\raggedright\arraybackslash}m{0.25\textwidth}@{}}
\multicolumn{7}{@{}l}{\textbf{Task generalization: box retrieval and placement}} \\[5pt]
\toprule
\textbf{Model} & \centering 0\,s (input) & \centering 6\,s & \centering 12\,s & \centering 22\,s & \centering 28\,s & \textbf{Judge summary} \\
\midrule
\gencell{\textbf{GT}\par\smallskip $G=5$, $C=0$} & \genframe{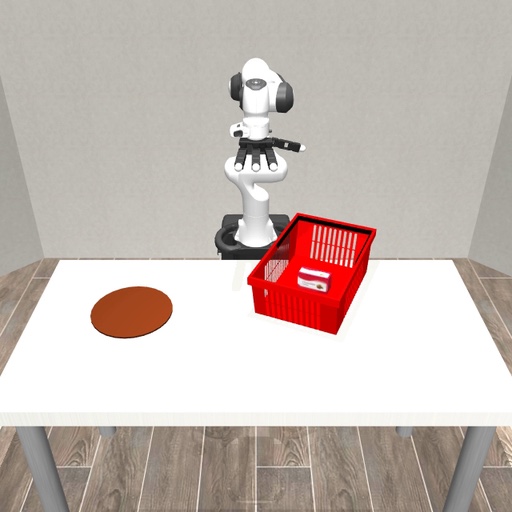} & \genframe{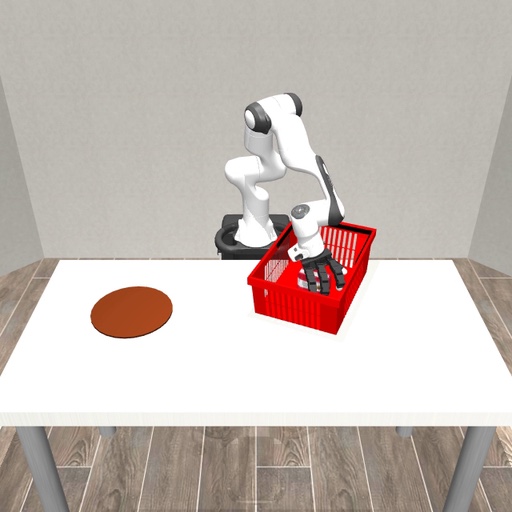} & \genframe{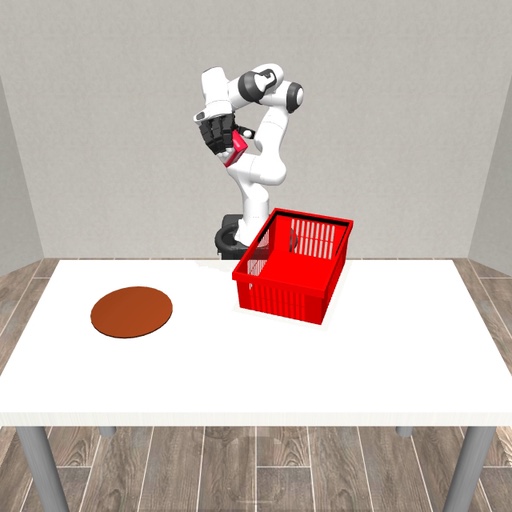} & \genframe{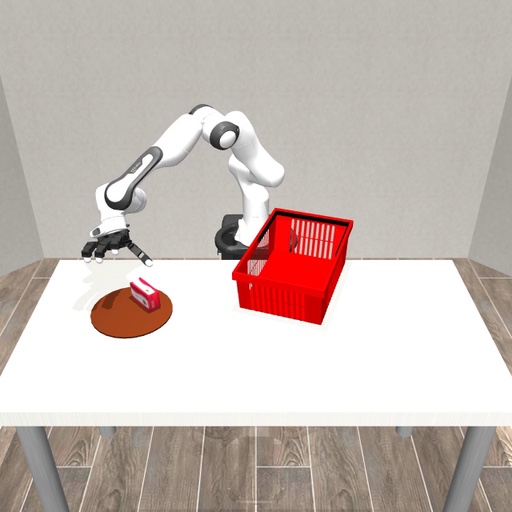} & \genframe{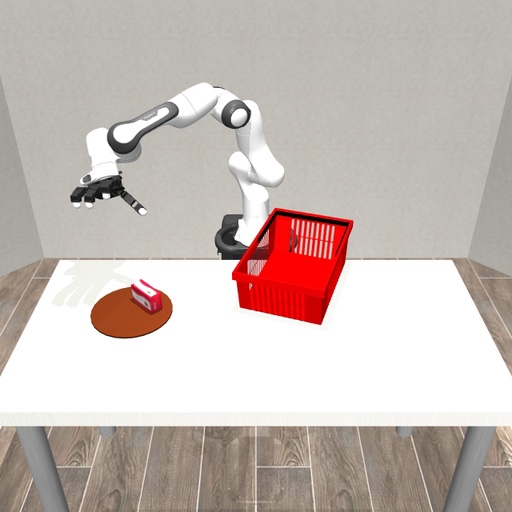} & \gencell{The box is extracted, transported, and released onto the plate. It remains there after the hand withdraws; scene entities remain coherent.} \\
\gencell{\textbf{Wan}\par\smallskip $G=1$, $C=1$} & \genframe{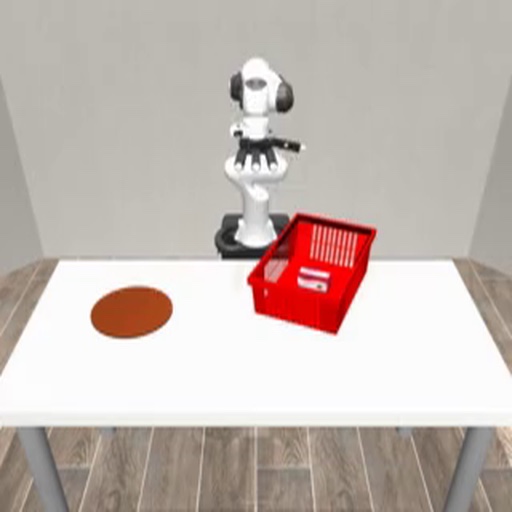} & \genframe{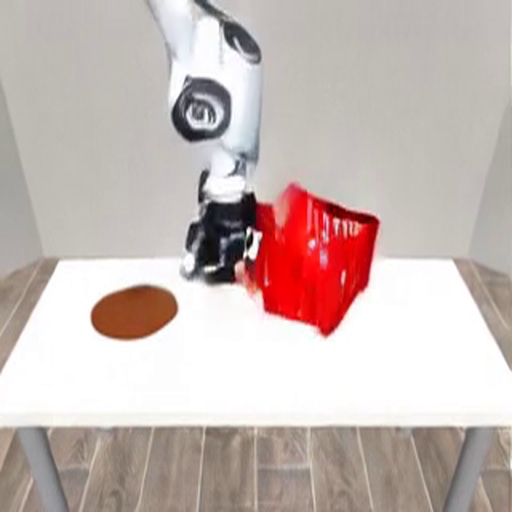} & \genframe{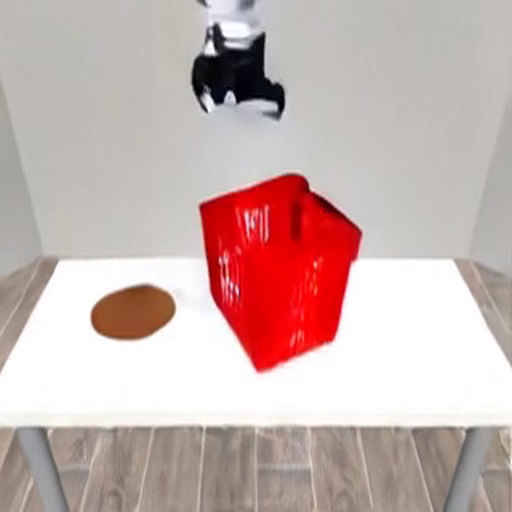} & \genframe{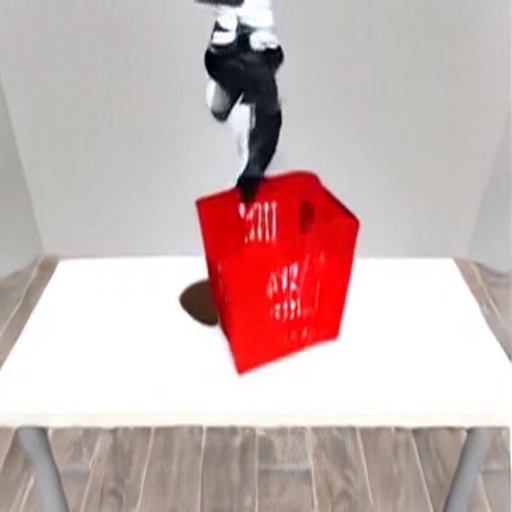} & \genframe{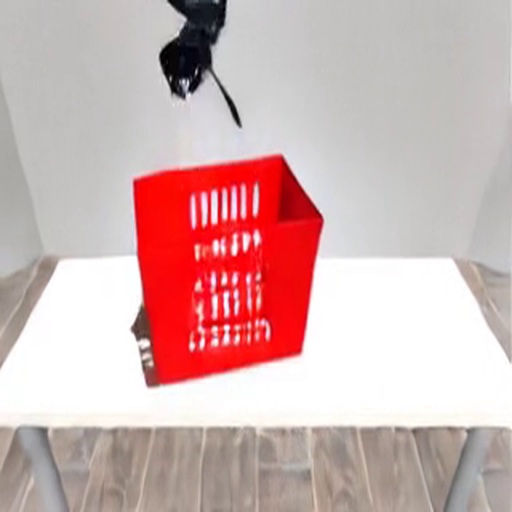} & \gencell{The robot primarily lifts the basket rather than extracting the box. No final placement is visible; basket walls and hand connections deform severely.} \\
\gencell{\textbf{IRASim}\par\smallskip $G=2$, $C=1$} & \genframe{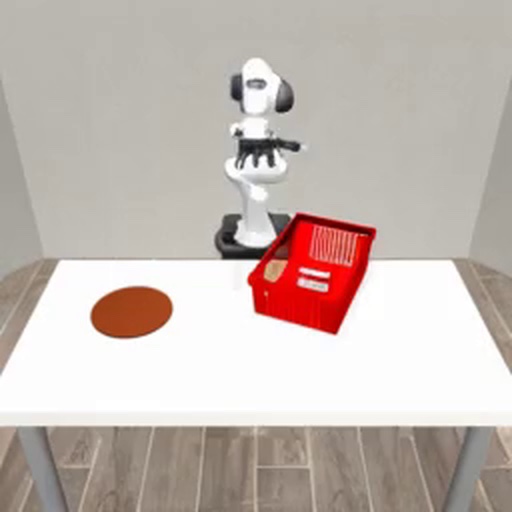} & \genframe{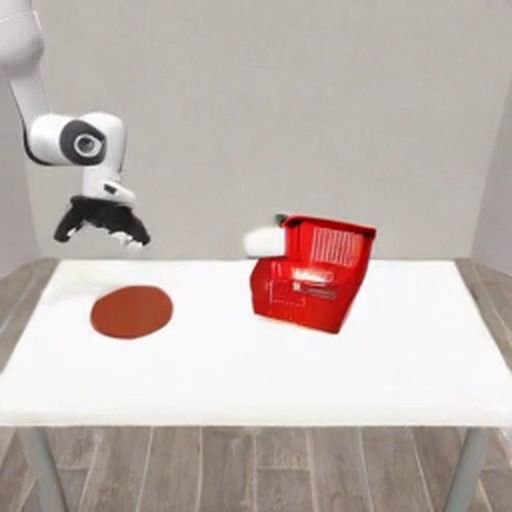} & \genframe{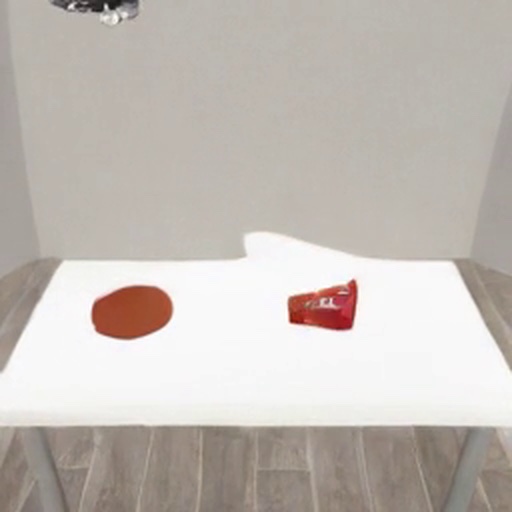} & \genframe{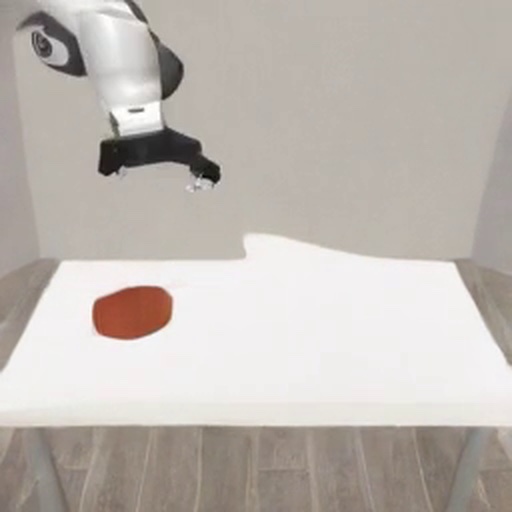} & \genframe{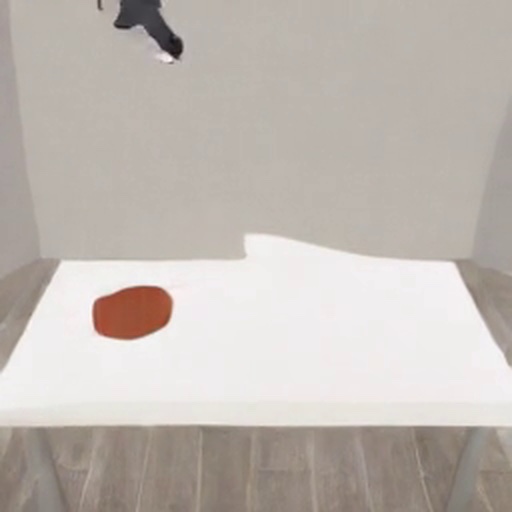} & \gencell{Task-directed manipulation is visible, but box extraction is unclear and the plate stays empty. The basket collapses and then disappears from the unobstructed table.} \\
\gencell{\textbf{Ctrl-World}\par\smallskip $G=2$, $C=0$} & \genframe{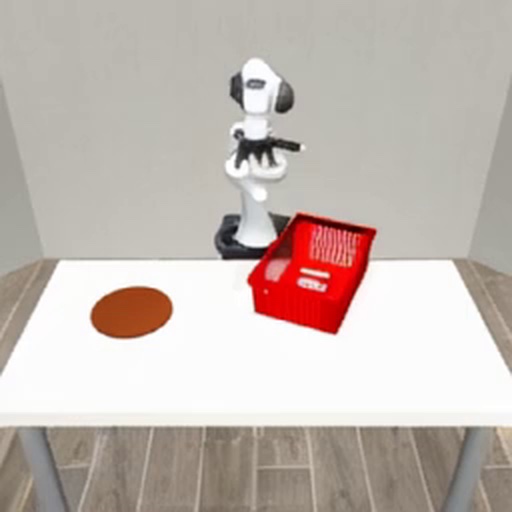} & \genframe{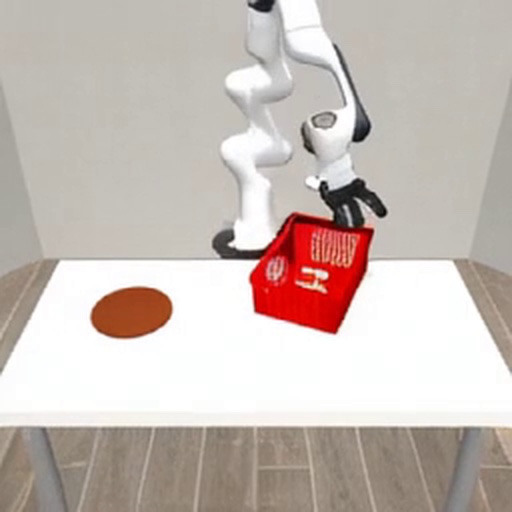} & \genframe{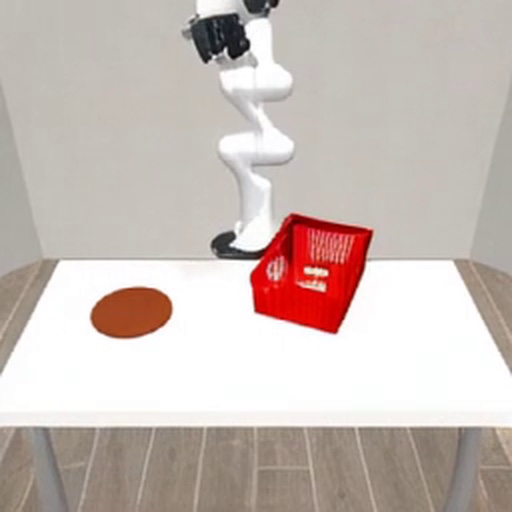} & \genframe{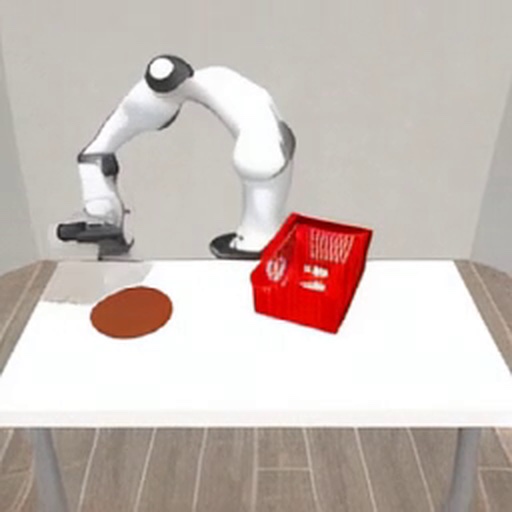} & \genframe{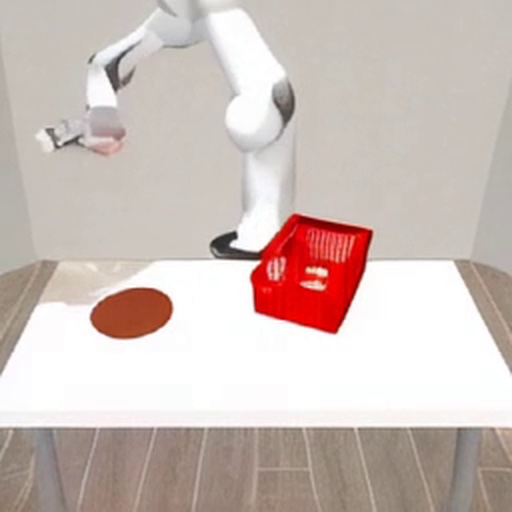} & \gencell{The hand attempts a grasp and moves toward the plate, but box extraction is unconfirmed and the plate stays empty. No severe collapse is observed.} \\
\gencell{\textbf{FastWAM}\par\smallskip $G=2$, $C=1$} & \genframe{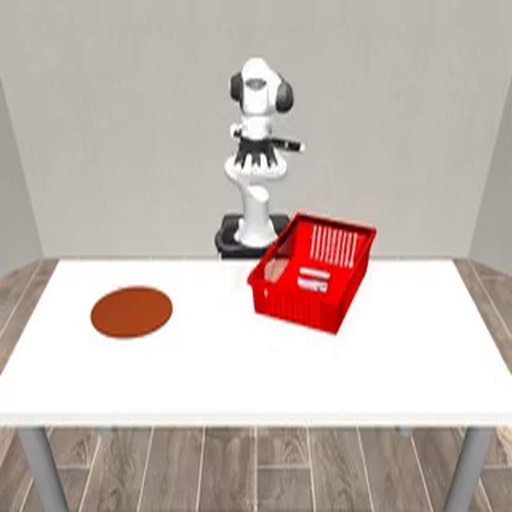} & \genframe{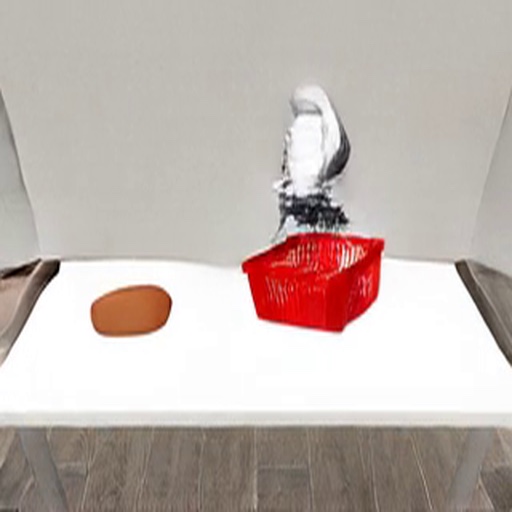} & \genframe{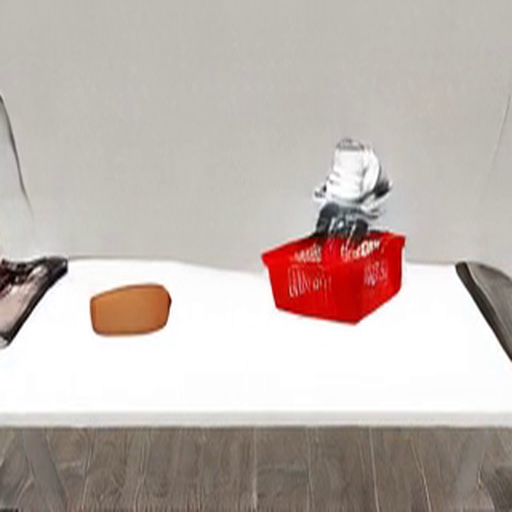} & \genframe{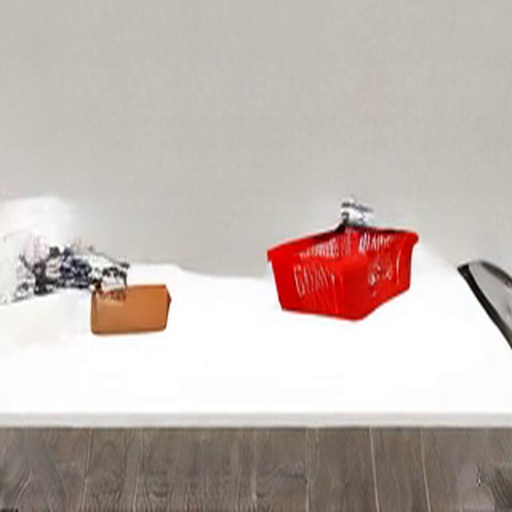} & \genframe{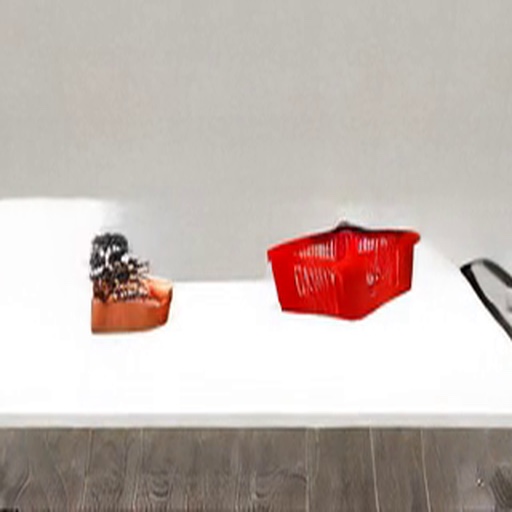} & \gencell{A grasp is attempted, but the box cannot be tracked to a supported placement. Wrist and hand structures become tangled, fragmented masses.} \\
\gencell{\textbf{Cosmos3}\par\smallskip $G=2$, $C=1$} & \genframe{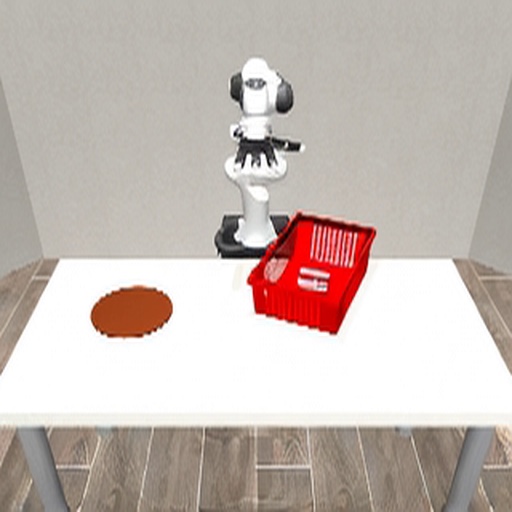} & \genframe{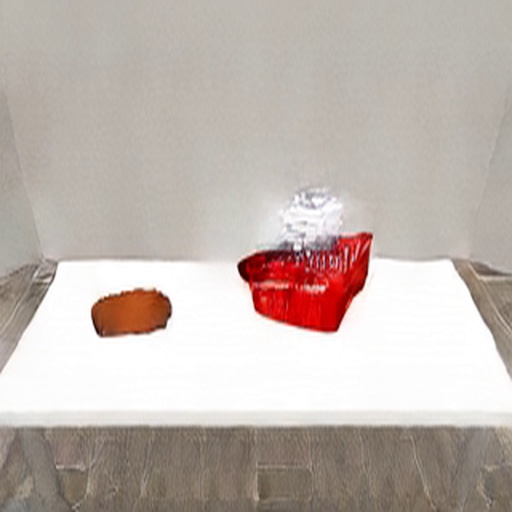} & \genframe{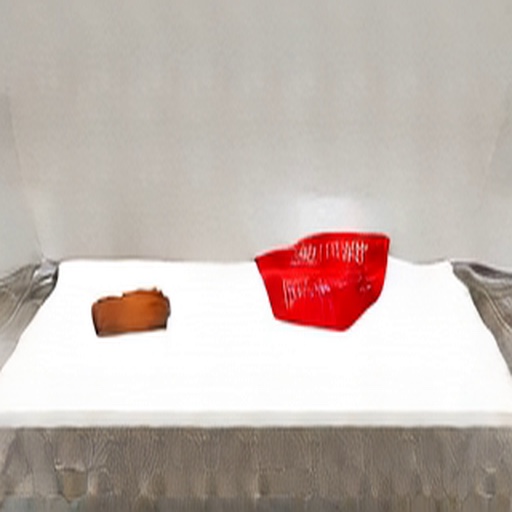} & \genframe{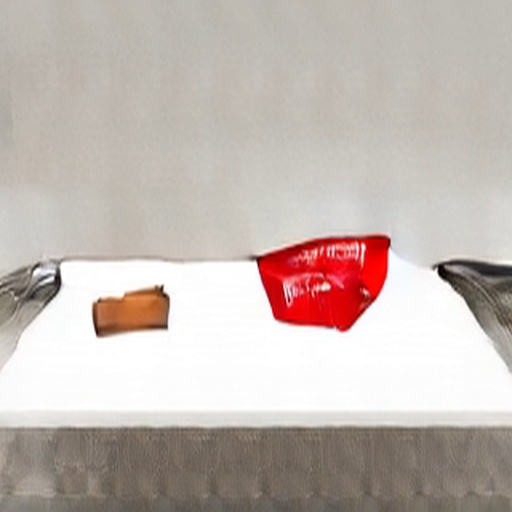} & \genframe{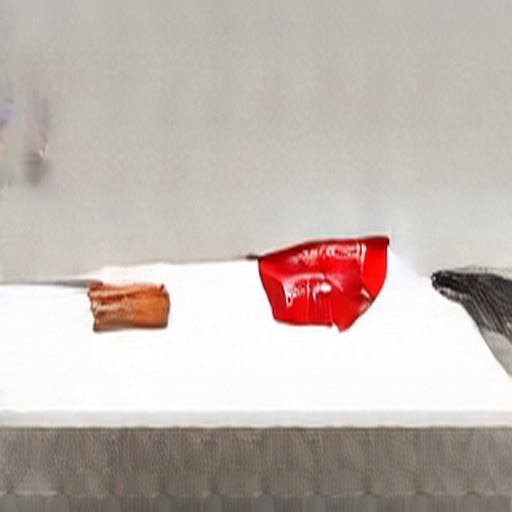} & \gencell{The hand attempts a grasp without confirmed box extraction or placement. Arm and hand structures fragment, while the basket collapses into a folded body.} \\
\gencell{\textbf{Mask2Real-\newline WM}\par\smallskip $G=2$, $C=0$} & \genframe{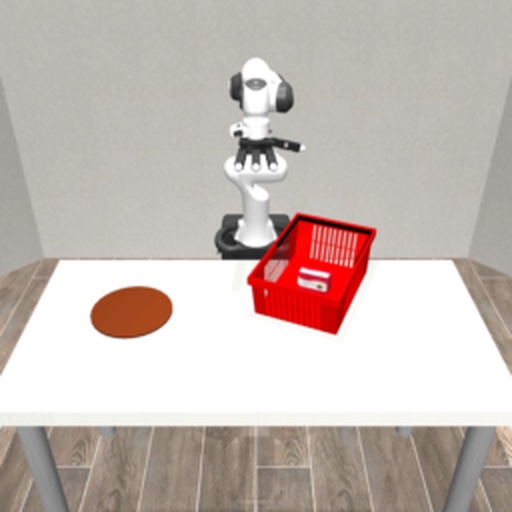} & \genframe{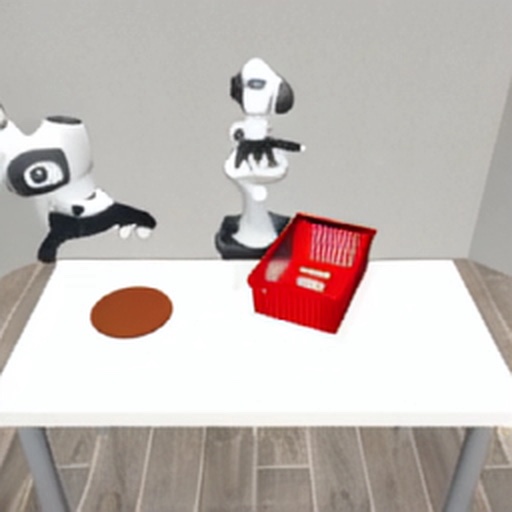} & \genframe{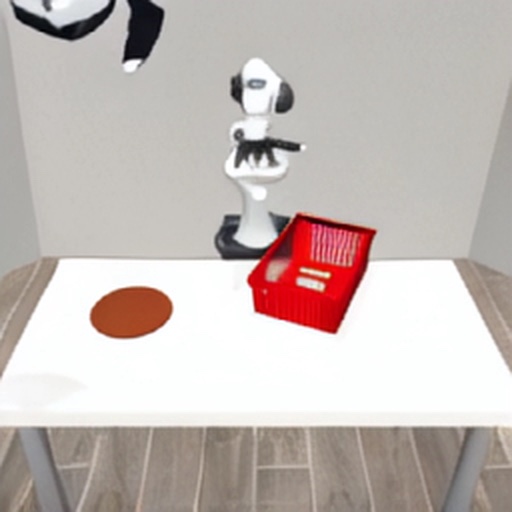} & \genframe{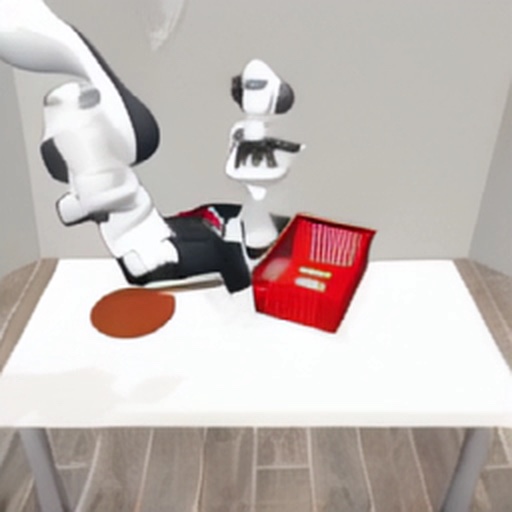} & \genframe{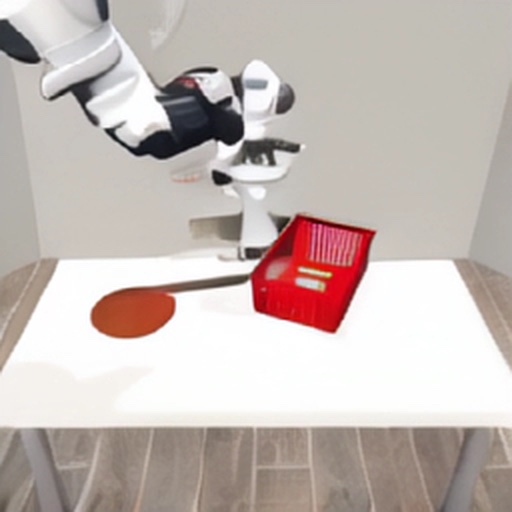} & \gencell{An attempted manipulation is visible, but grasping, extraction, and final placement remain unconfirmed. The arm, hand, and basket retain coherent structures.} \\
\bottomrule
\end{tabular}
\endgroup
\caption{\textbf{Retrieving the box from the bucket and placing it on the plate.} Ctrl-World and Mask2Real-WM retain coherent imagery but do not establish successful placement. $G$: task-completion score; $C$: visual collapse. Reasons summarize the original judgments; scores refer to the full judged sequence.}
\label{fig:gen_task}
\end{figure}

\end{document}